\documentclass[10pt,twocolumn,letterpaper]{article}

\usepackage{cvpr}              

\usepackage{cuted} 

\usepackage{currfile} 

\usepackage{caption} 

\definecolor{cvprblue}{rgb}{0.21,0.49,0.74}
\usepackage[pagebackref,breaklinks,colorlinks,allcolors=cvprblue]{hyperref}

\usepackage{xcolor}
\definecolor{ourblue}{HTML}{0072B2}
\definecolor{ourorange}{HTML}{D55E00}
\definecolor{ourgreen}{HTML}{009E73}

\usepackage{colortbl}
\definecolor{vtUns}{HTML}{F0F0F0}
\definecolor{vtWeak}{HTML}{E7ECF6}
\definecolor{vtFine}{HTML}{F8ECDD}
\definecolor{vtTF}{HTML}{EEF6EA}
\definecolor{vtFam}{HTML}{DCEED3}
\definecolor{vtVIDE}{HTML}{C4E6CF}
\definecolor{vtAcc}{HTML}{0B6E69}
\newcommand{\vtours}{\textcolor{vtAcc}{\,[ours]}}
\newcommand{\vtbase}{\textcolor{black!55}{\,[baseline]}}

\usepackage{algorithm}
\usepackage{algpseudocode}

\def\paperID{36951} 
\def\confName{CVPR}
\def\confYear{2027}

\title{VIDE : Training-Free Video Anomaly Detection \\ via Visual Intrinsic Distribution Estimation}

\author{Kevin D. Nguetche ,  \qquad Atif Belal,  \qquad Lilian Hollard,  \qquad Eric Granger,  \qquad Marco Pedersoli\\
LIVIA, Dept. of Systems Engineering, ETS Montreal, Canada \\
International Laboratory on Learning Systems (ILLS)\\
}

\begin{document}
\maketitle
\begin{abstract}
In embedding-based video anomaly detection (VAD), normal clips can have high similarity to anomalous descriptions, an anomaly bias that leads to false positives. Methods such as PRISM mitigate this bias by measuring deviations from a normality reference along anomaly directions. Both the normality reference and the anomaly directions are derived from text and shared across videos, although normal activity and the appearance of anomalies vary across scenes. The resulting scores are also sensitive to prompt selection.
To address these limitations, we propose Visual Intrinsic Distribution Estimation (VIDE), a training-free framework that adapts text-derived scoring to each test video using frozen clip embeddings. VIDE estimates normal scene content from clips with low initial anomaly scores and uses this estimate to partially correct the textual reference. It then rescales category responses, penalizes clips with strong responses across categories, and refines the anomaly directions using high-scoring clips. VIDE averages scores across prompt views to reduce sensitivity to prompt selection and fuses scores from two frozen encoders.
With both encoders, VIDE improves AUC over the best PRISM result by $1.5$, $1.9$, and $3.5$ points on UCF-Crime, XD-Violence, and MSAD, respectively, and AP by $6.5$ points on XD-Violence and $9.3$ points on MSAD. VIDE yields the highest AUC among the compared training-free methods on all three benchmarks and the highest reported AUC among all compared methods on XD-Violence and MSAD. Our code is included in the supplementary material and will be made publicly available.
\end{abstract}

\section{Introduction}
\label{sec:intro}



\begin{figure}[t]
  \centering
  \includegraphics[width=\linewidth]{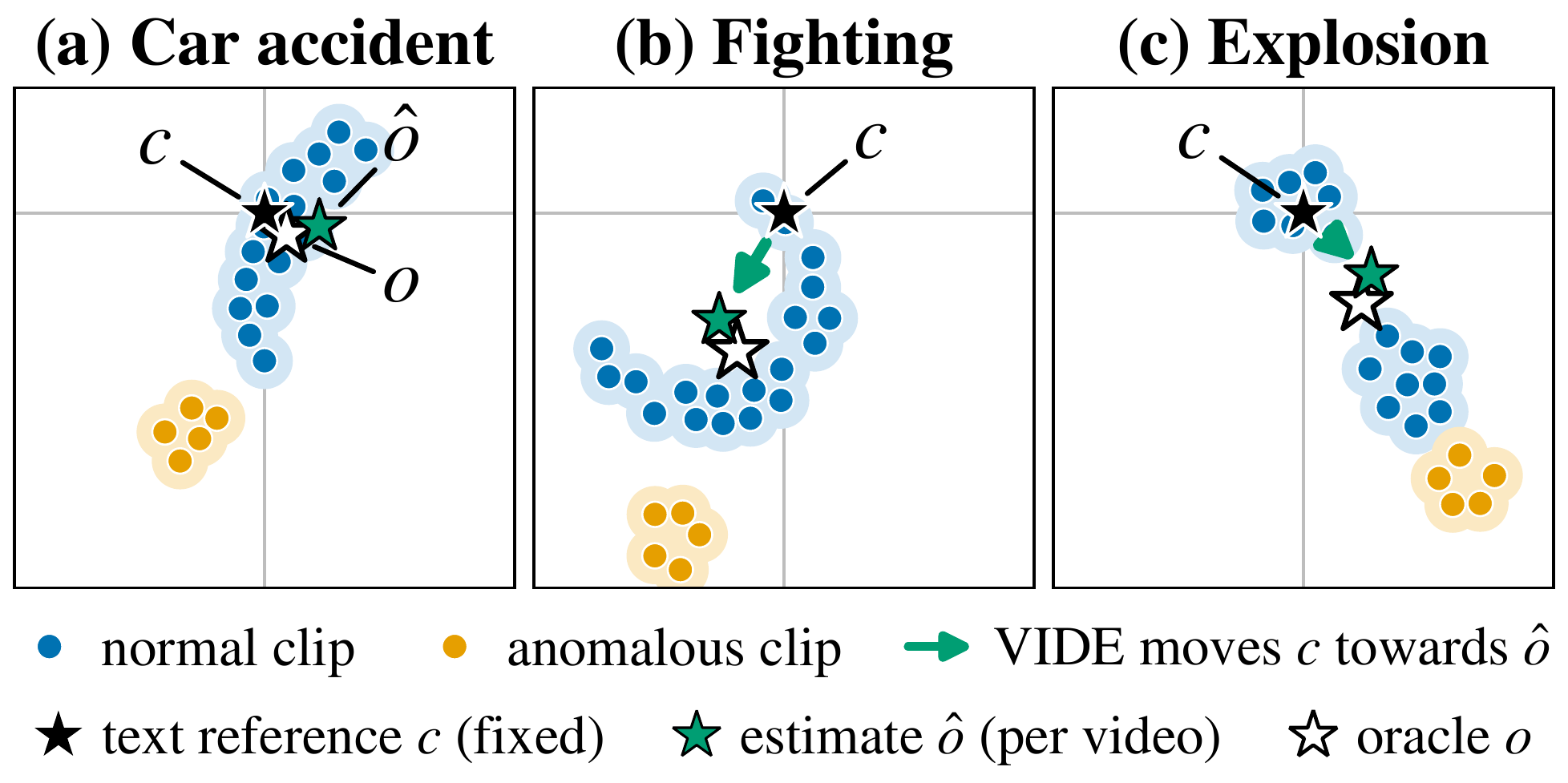}
  \vspace{-1.5em}
  \caption{\textbf{t-SNE illustration of the textual and visual references (Qwen3-VL-Embedding-2B, XD-Violence videos).} The textual reference $c$ of PRISM, the mean of the normal prompts, is applied to every video without looking at it: it works when the video matches the prompts (a), where it is close to the annotated normal centre $o$ (oracle), but lies far from it otherwise (b, c). VIDE estimates $\hat{o}$ from the video itself, without annotation or training; $\hat{o}$ stays close to $o$ in all three cases. Where needed, VIDE partially corrects $c$ towards $\hat{o}$ (green arrow).}
  \vspace{.em}
  \label{fig:teaser}
\end{figure}

Video anomaly detection (VAD) aims to temporally localize events that deviate from normal behavior in a scene~\cite{sultani2018real}. Training-based methods learn to distinguish normal from anomalous activity using normal-only videos~\cite{hasan2016learning,liu2018future,gong2019memorizing,park2020learning}, video-level annotations~\cite{sultani2018real,wan2021weakly,joo2023clip}, or pseudo-labels derived from unlabeled videos~\cite{al2024coarse}. Recent approaches incorporate pretrained vision-language models to improve detection performance~\cite{wu2024vadclip,wu2024open,zhang2025holmes,ye2025vera,huang2025ex}, but remain dependent on task-specific training. These detectors often generalize poorly beyond their training domains, requiring additional data collection and retraining to maintain performance in new environments~\cite{zanella2024harnessing}. These limitations have motivated recent work on training-free VAD~\cite{zanella2024harnessing, shao2025eventvad, chenprism}, which exploits pretrained models to detect anomalies without additional model training.

Existing training-free VAD methods can be broadly categorized into generative and embedding-based approaches. Generative methods infer anomaly scores from generated video descriptions or through direct multimodal prompting~\cite{zanella2024harnessing,shao2025eventvad,li2026vadtree,Zou2025UnlockingVM}. However, repeated text generation incurs substantial computational overhead, and errors in generated descriptions can affect detection accuracy. Embedding-based methods compute anomaly scores directly from visual and textual representations obtained from frozen encoders~\cite{lee2025flashback,chenprism}. By avoiding text generation during inference, these methods provide an efficient alternative for training-free VAD.

Despite their efficiency, embedding-based methods are susceptible to anomaly bias, whereby normal clips exhibit high similarity to anomalous descriptions~\cite{lee2025flashback,chenprism}. PRISM mitigates this bias by subtracting a normality reference and constructing decorrelated anomaly directions from text statistics~\cite{chenprism}. However, these statistics are computed from text alone and applied unchanged to every test video, although normal activity varies across scenes. 
As \cref{fig:teaser} illustrates, the same textual reference can lie close to the normal clips of one video and far from those of another. A reference that does not represent the normal activity of a given scene can raise scores across its normal clips, and poorly aligned anomaly directions can weaken the contrast between an event and its surroundings. These effects can increase false positives and impair temporal localization, even when the textual descriptions correctly specify the events of interest. Concurrent work uses a visual normality anchor to account for scene context~\cite{Yin2026NOVANM}, but adapting the reference alone does not ensure that the anomaly directions align with the visual content of the video. 

Given these considerations, we ask: \textit{can the normality reference and anomaly directions be refined using the test video itself, without labels or gradient updates, while preserving the efficiency of embedding-based methods?} To answer this question, we introduce VIDE (Visual Intrinsic Distribution Estimation), a training-free framework that adapts anomaly scoring to each test video. VIDE first estimates normal scene content by assigning greater weight to clips with low initial anomaly scores. This visual estimate guides a partial correction of the textual reference, preserving the semantic information provided by text (\cref{fig:teaser}). VIDE then calibrates anomaly responses to a common scale, penalizes clips with strong responses across multiple categories, and refines the anomaly directions using high-scoring clips. Finally, it averages scores across several prompt subsets to reduce sensitivity to prompt wording and fuses the scores of two frozen encoders.

\noindent\textbf{Our main contributions are summarized as follows:}
\begin{itemize}
    \item We analyze the limitations of text-derived anomaly scoring, identifying a mismatch between the normality reference and normal scene content, misaligned anomaly directions, and sensitivity to prompt selection.
    \item We propose VIDE, a training-free framework that combines textual statistics with visual estimates from each test video to refine the normality reference and anomaly directions. It operates on precomputed embeddings without labels or gradient updates.
    \item VIDE achieves the best AUC among the compared training-free methods on UCF-Crime, XD-Violence, and MSAD, and the best AUC among all compared methods on XD-Violence and MSAD.
\end{itemize}

\vspace{-.5em}
\section{Related Work}
\label{sec:related_work}

\begin{figure*}[t]
  \centering
  \includegraphics[width=\textwidth]{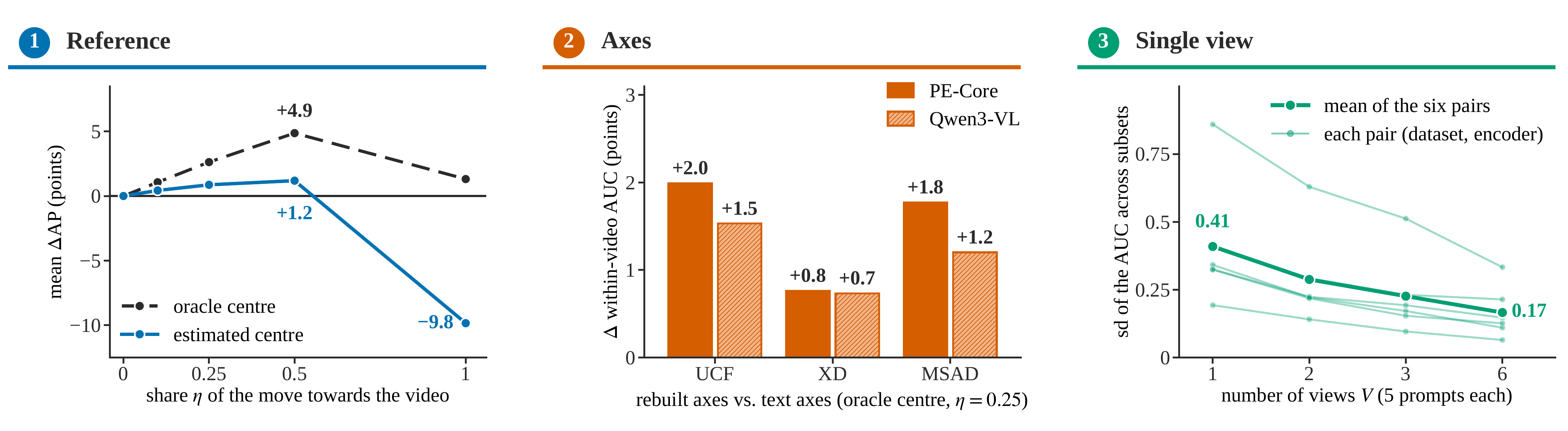}
  \vspace{-2.em}
  \caption{\textbf{Limitations of text-derived scoring}. \textbf{(1)~Reference:} mean change of AP over the six (3 datasets, 2 encoders) pairs when the reference moves by a share $\eta$ towards a video centre, estimated without labels (estimated centre) or computed on the normal clips (oracle centre). \textbf{(2) Axes:} gain in within-video AUC from recomputing the anomaly directions, with the reference fixed at $c_\eta(o)$ and $\eta=0.25$ in both configurations. \textbf{(3)~Single view:} standard deviation of the AUC over ensembles of $V$ subsets of 5~prompts; thin: each of the six pairs, thick: their mean.}
  \label{fig:diag}
\end{figure*}

\noindent\textbf{Training-based VAD.} One-class methods model normal activity through frame reconstruction~\cite{hasan2016learning,gong2019memorizing}, frame prediction~\cite{liu2018future,park2020learning}, or discriminative subspaces~\cite{wang2019gods}. Unsupervised methods derive pseudo-labels through generative cooperative learning~\cite{zaheer2022generative} or clustering~\cite{al2024coarse}, and DyAnNet uses scene dynamics to guide self-training~\cite{thakare2023dyannet}. Weakly supervised methods use multiple instance learning to train segment-level anomaly scorers from video-level labels~\cite{sultani2018real,tian2021weakly,zhou2023dual}. Recent methods incorporate poly-modal inputs~\cite{majhi2025just}, mixtures of experts~\cite{amicantonio2025mixture}, or disentangled semantic alignment~\cite{yin2026learning}. CLIP-TSA trains a detector on frozen CLIP features~\cite{joo2023clip}, and VadCLIP incorporates text into anomaly scoring through vision-language alignment~\cite{wu2024vadclip}. Open-vocabulary approaches use language priors to extend detection to anomaly categories absent from training~\cite{wu2024open,li2025anomize,liu2026language}. For explainable VAD, some methods train vision-language models~\cite{huang2025ex} or fine-tune multimodal models~\cite{Zhang2024HolmesVADTU,zhang2025holmes,Zhu2025VAUR1AV}, and others learn questions that guide a frozen model~\cite{ye2025vera}. Although effective in their training domains, these methods may require extra video data and retraining to maintain detection performance in new environments.

\noindent\textbf{Training-free VAD with generative models.} These methods query frozen generative models at inference time. LAVAD uses a language model to summarize frame captions over time and estimate anomaly scores~\cite{zanella2024harnessing}. MCANet and SUVAD also score generated descriptions, with MCANet incorporating captions from both visual and audio inputs~\cite{dev2024mcanet,gao2025suvad}. EventVAD segments videos into events for scoring~\cite{shao2025eventvad}, and VADTree organizes events into a hierarchy to score them at multiple temporal scales~\cite{li2026vadtree}. ASK-HINT queries a frozen vision-language model with questions about fine-grained actions~\cite{Zou2025UnlockingVM}, and HiProbe-VAD derives anomaly scores from intermediate hidden states of a multimodal language model~\cite{Cai2025HiProbeVADVA}. Several of these methods provide textual explanations. Running generative models on each test video can be computationally expensive, and errors in generated captions can propagate to anomaly scores in caption-based pipelines.

\noindent\textbf{Training-free VAD with embeddings.} These methods score video clips by comparing their embeddings with normal and anomalous text descriptions, an approach also used in image anomaly detection~\cite{Jeong2023WinCLIPZA,Zhou2023AnomalyCLIPOP}. Flashback~\cite{lee2025flashback} builds a bank of normal and anomalous captions offline and downweights anomalous caption embeddings before retrieval. PRISM~\cite{chenprism} uses the mean embedding of the normal prompts as its reference and whitens anomaly directions using the covariance of both normal and anomalous prompt embeddings. These statistics are shared across videos and can be poorly matched to the visual content of individual scenes. NOVA~\cite{Yin2026NOVANM}, a concurrent work, excludes ambiguous action terms during normal prompt generation and adds a visual anchor, estimated from the initial frames, to the set of normal embeddings used for scoring. VIDE also uses an anchor estimated from the initial clips, but it shifts the scoring reference instead of adding the anchor to the prompts. It extends PRISM's scoring head by partially correcting the textual reference using a normality estimate from clips across the video, refines the anomaly directions using high-scoring clips, and calibrates their responses (\hyperref[sec:supp_nova]{Appendix}).

\noindent\textbf{Statistics of embedding spaces.} Text embeddings can be anisotropic~\cite{gao2019representation}, and a dominant component shared by visual and textual embeddings can inflate the similarity between normal clips and anomalous descriptions~\cite{chenprism}. The modality gap between visual and textual embeddings~\cite{Liang2022MindTG} can also make a text-derived reference less representative of normal video clips. Hubness describes the tendency of some embeddings to appear among the nearest neighbors of many queries~\cite{Radovanovi2010HubsIS,Bogolin2021CrossMR}. This provides an analogy for clips that respond strongly to multiple anomaly categories. Vision-language models are sensitive to prompt wording~\cite{Radford2021LearningTV,Zhou2021LearningTP}. VIDE partially corrects the normality reference using visual evidence, penalizes broad responses across categories, and averages scores over prompt subsets. Using estimates from each test video to correct text-derived scoring statistics parallels the re-estimation of normalization statistics from unlabeled target data in domain adaptation~\cite{Li2016RevisitingBN,Schneider2020ImprovingRA}.

\vspace{-.8em}

\section{Method}
\label{sec:method}

\subsection{Setting and notation}
\label{sec:setting}

We divide each test video into $T$ clips. A frozen vision-language encoder maps clip $t$ to a unit-norm embedding $x_t \in \mathbb{R}^D$. The text branch of the same encoder embeds a prompt bank containing descriptions of normal activity and descriptions for each of $M$ anomaly categories or directions.

Let $\mu_n$ denote the mean embedding of the normal prompts and $\mu_j$ the mean embedding of the prompts for anomaly category $j$. We compute the covariance $\Sigma$ over all normal and anomalous prompt embeddings. For a regularisation parameter $\lambda>0$, we define the whitening matrix $W$ and the soft projector $P$ as
\begin{equation}
  W = (\Sigma + \lambda I)^{-1/2},
  \qquad
  P = \Sigma(\Sigma + \lambda I)^{-1},
\end{equation}
where $I$ is the identity matrix. The projector $P$ attenuates components along directions with low variance across prompt embeddings. Given encoder and prompt set, these text statistics are computed once and reused across videos.

VIDE builds on the scoring head of PRISM~\cite{chenprism}, which uses $c=\mu_n$ as the normality reference and constructs one anomaly direction (axis) per category, $a_j=\operatorname{unit}\big(W(\mu_j-c)\big)$, where $\operatorname{unit}(z)=z/\lVert z\rVert_2$. The logit for clip $t$ and category $j$ is $L_{t,j}=(x_t-c)^\top a_j$. Given the logit vector $L_t=(L_{t,1},\ldots,L_{t,M})$, the clip score is $s_t=\sum_{j=1}^{M}\pi_{t,j}L_{t,j}$, where $\pi_t=\operatorname{softmax}(L_t/\tau)$ and $\tau>0$ is the temperature. PRISM smooths the resulting scores over time with a Gaussian filter of standard deviation $\sigma$. VIDE extends this head using visual statistics estimated independently from the unlabelled clips of each test video.

\subsection{Limitations of text-derived scoring}
\label{sec:discrepancies}

Let $o$ denote the unit-normalised mean embedding of the normal clips in a video. The logit for clip $t$ and anomaly category $j$ can be written as
\begin{equation}
  L_{t,j} = (x_t - o)^{\top} a_j + (o - c)^{\top} a_j .
  \label{eq:decomposition}
\end{equation}
The first term measures how much the clip deviates from the video's normal content along $a_j$. The second is a video- and category-specific offset shared by all clips. Both the reference $c$ and the directions $a_j$ depend on the prompt set.

We compute the oracle centre $o$ from annotated normal clips for diagnostic comparisons only. VIDE uses the label-free estimate $\hat{o}$, which assigns greater weight to clips with low initial anomaly scores (\cref{sec:reference}). We evaluate all six combinations of three datasets (UCF-Crime, XD-Violence, and MSAD) and two encoders (PE-Core and Qwen3-VL-Embedding) using the published PRISM parameters. We compute between-video AUC by assigning each clip its video's mean score. For within-video AUC, we subtract that mean from each clip score and compute a single AUC over all centered scores.

\noindent\textbf{Normality reference.}
Mean video scores, which include the offset in \cref{eq:decomposition}, carry a strong detection signal. Between-video AUC ranges from 83 to 92\%, and within-video AUC ranges from 58 to 72\%. Across videos, the standard deviation of the mean score of normal clips is 1.0--2.6 times the average score difference between anomalous and normal clips. To test reference correction, we set $c_\eta(y)=\mu_n+\eta P(y-\mu_n)$, where $y$ is either the oracle centre $o$ or the estimated centre $\hat{o}$ (\cref{fig:diag}, panel~1). At $\eta=0.5$, the oracle and estimated centres improve mean AP by 4.9 and 1.2 points, respectively. AUC changes range from $+0.3$ to $+2.0$ points with the oracle centre and from $-0.8$ to $+1.6$ points with the estimated centre. At $\eta=1$, the estimated centre reduces mean AP by 9.8 points. The oracle centre yields a mean AP gain of 1.3 points and an AUC drop of 1.2--7.4 points across the six pairs. These results motivate a partial correction of the textual reference.

\noindent\textbf{Anomaly directions.}
We compare each text-derived direction with the mean direction of the corresponding anomalous clips, measuring each clip's direction from the normal centre of its video. Their cosine similarity ranges from 0.16 to 0.40. The text direction for the correct category is the most closely aligned for 45--83\% of the anomaly categories, compared with chance rates of $1/M$ (8--17\%). To isolate the effect of rebuilding the directions, we use the same corrected reference $c_\eta(o)$ with $\eta=0.25$ and compare the original directions with those recomputed as $\operatorname{unit}\big(W(\mu_j-c_\eta(o))\big)$ (\cref{fig:diag}, panel~2). Recomputing the directions improves within-video AUC by 0.7--2.0 points and AP by 0.8--4.3 points across all six pairs. Changes in overall AUC range from $-0.6$ to $+0.6$ points. With the label-free centre, the gain in within-video AUC is at most 0.6 point. The dispersion of the responses also differs by up to a factor of two across directions. These findings motivate refining the directions using visual evidence and expressing their responses in units of the text separation of each category, rather than equalizing their dispersion.

\noindent\textbf{Prompt sensitivity.}
We draw random subsets of five prompts per category, independently and without replacement within a subset, so that different subsets may share prompts. We compute AUC after averaging scores over $V$ subsets (\cref{fig:diag}, panel~3). With a single subset, the standard deviation of AUC across draws ranges from 0.19 to 0.86 points, with a mean of 0.41. The largest deviations from the full prompt bank occur on XD-Violence with Qwen3-VL-Embedding and range from $-2.2$ to $+1.2$ points. With six subsets, the mean standard deviation falls to 0.17 points, and the mean AUC remains within 0.6 points of the full-bank result. Averaging scores across subsets therefore reduces AUC variability due to prompt selection.

\begin{figure}[t]
  \centering
  \includegraphics[width=\columnwidth]{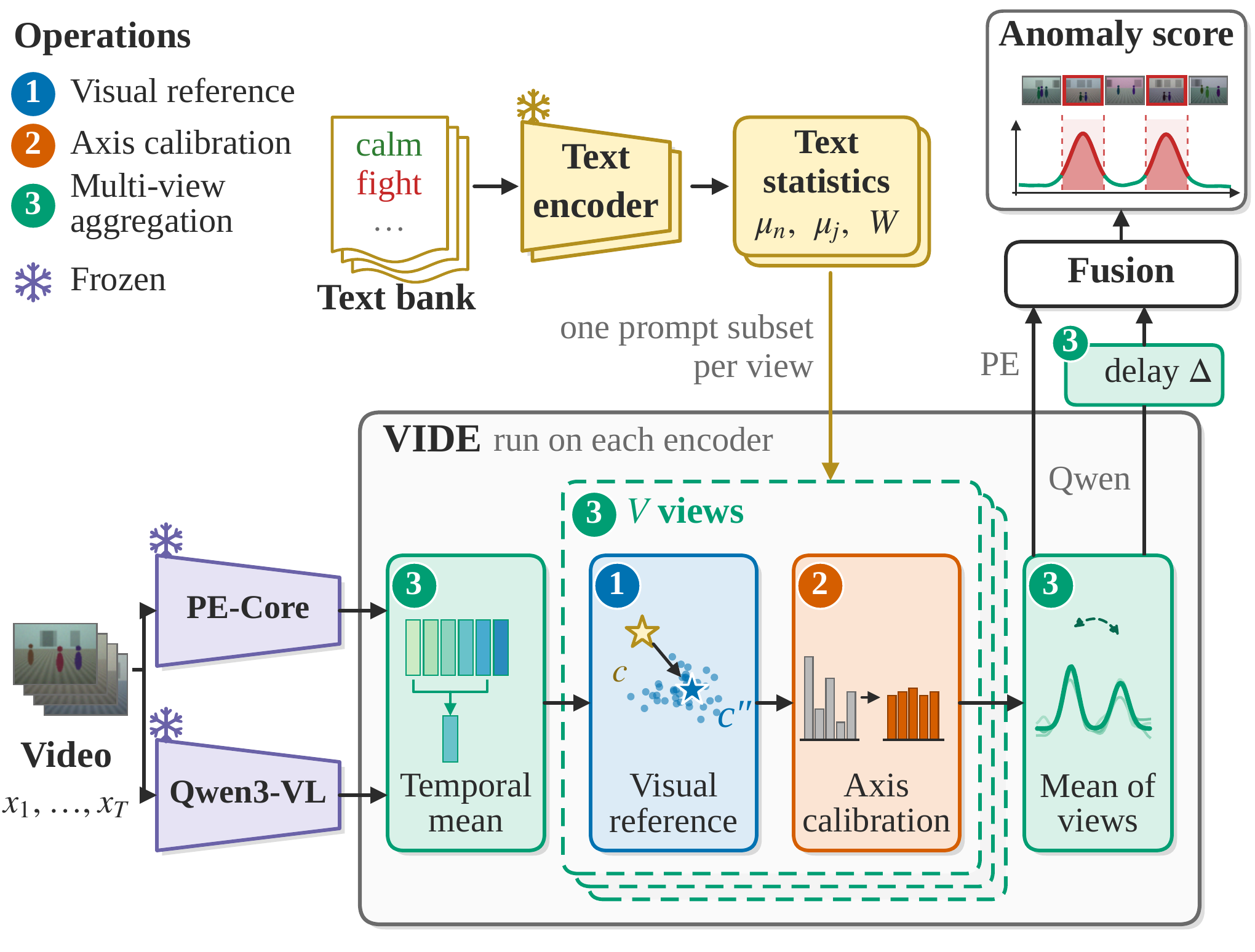}
  \vspace{-2.3em}
  \caption{\textbf{Overview of VIDE.} The text bank gives the text statistics
  $(\mu_n,\mu_j,W, P)$ once. On each frozen encoder, VIDE averages neighbouring
  clips, then runs $V$ views, each with its own set of prompts. In each view, Operation~\textcolor{ourblue}{1} uses the corrected reference $c'$ to construct the anomaly directions and the scoring reference $c''$ to compute the logits. Operation~\textcolor{ourorange}{2} calibrates the category responses.
  Operation~\textcolor{ourgreen}{3} averages the views and delays the Qwen score ($\Delta$). The PE score and the
  delayed Qwen score are fused into the anomaly score.}
  \label{fig:overview}
\end{figure}

\vspace{-.5em}

\subsection{Overview of VIDE}
\label{sec:overview}

VIDE adapts PRISM's scoring head to each test video using embeddings from frozen encoders. For each encoder, it scores the video under $V$ prompt views. The first view uses the full prompt bank, and each additional view uses a subset of the anomaly prompts. Operation~1 (\cref{sec:reference}) and Operation~2 (\cref{sec:calibration}) are applied separately in each view. Operation~3 (\cref{sec:views}) smooths the clip embeddings before scoring, combines the resulting scores, and delays the Qwen score (\cref{fig:overview}).

\noindent\textbf{Score aggregation and delay.}
Before scoring, VIDE averages neighbouring clip embeddings over a window of $T_w$ clips and normalises the result, denoted by $\tilde{x}_t$. Each view $v$ yields clip scores $s^{(v)}_t$ (\cref{alg:calibration}). VIDE averages the scores from all views and can combine each clip's score with those of similar clips elsewhere in the same video, giving
\begin{align}
  \bar{s}_t
    &= \frac{1}{V}\sum_{v=1}^{V} s^{(v)}_t,
  \nonumber\\
  \tilde{s}_t
    &= (1-\omega_c)\bar{s}_t
       + \frac{\omega_c}{\kappa}\sum_{u\in\mathcal{N}_t}\bar{s}_u.
  \label{eq:final}
\end{align}
The set $\mathcal{N}_t$ contains the $\kappa$ clips most similar to clip $t$ among those satisfying $|t-u|>\delta$. When $\kappa=0$, we set $\tilde{s}_t=\bar{s}_t$. A Qwen embedding covers a window of $7$ consecutive clips ($112$ frames), so the Qwen score reacts to an event a few clips before the PE score. For Qwen, the score sequence $\tilde{s}$ is therefore delayed by $\Delta$ clips, and its first score fills the start. The two scores added at a clip then describe the same instant. $\Delta$ is selected by leave-one-dataset-out with the head and the fusion (\hyperref[sec:supp_hyper]{Appendix}). This delay is not specific to VIDE: on MSAD, it also raises the AUC of PRISM (Qwen) by $0.65$ points and its AP by $3.10$ points, against $0.68$ and $3.46$ for VIDE (\hyperref[sec:supp_delay]{Appendix}).

For a single encoder, setting $\eta=\omega_s=p=\beta=\omega_m=0$, $T_w=1$, $\kappa=0$, and $V=1$ (and $\Delta=0$ for Qwen) recovers the PRISM scoring head~\cite{chenprism}. Scoring the views scales linearly with $T$ and $V$, and the optional pairwise neighbour search scales quadratically with $T$. For the final prediction, VIDE smooths the PE-Core and Qwen3-VL-Embedding score sequences and combines them in a weighted sum, with weight $\phi$ on the delayed Qwen scores.

\vspace{-.5em}
\subsection{Visual reference estimation}
\label{sec:reference}

The results in \cref{sec:discrepancies} motivate a partial correction of the textual reference toward the normal content of each video. We apply \cref{alg:reference} independently to each prompt view using its text statistics and omit the view index below.

\begin{algorithm}[t]
\caption{Visual reference estimation for one prompt view.}
\label{alg:reference}
\fontsize{8}{9}\selectfont
\begin{algorithmic}[1]
\Require Clip embeddings $\tilde{x}_1,\dots,\tilde{x}_T$, text statistics $\mu_n$,
$\mu_j$, $W$, $P$, temperature $\tau$, and parameters $\gamma$, $\eta$, $K$,
$\gamma_r$, $\omega_s$
\Ensure Anomaly directions $a'_j$ and references $c'$ and $c''$
\State $s^0 \leftarrow \operatorname{head}\big((\tilde{x}_t-\mu_n)^{\top}
  \operatorname{unit}(W(\mu_j-\mu_n))\big)$
\State $w_t \leftarrow \exp\!\big(-(s^0_t-\min_u s^0_u)/\gamma\big)$, then
  normalise so that $\sum_t w_t=1$
\State $\hat{o} \leftarrow \operatorname{unit}\big(\sum_t w_t\,\tilde{x}_t\big)$
\State $c' \leftarrow \mu_n+\eta P(\hat{o}-\mu_n)$
\State $a'_j \leftarrow \operatorname{unit}\big(W(\mu_j-c')\big)$, \quad
  $s^1 \leftarrow \operatorname{head}\big((\tilde{x}_t-c')^{\top}a'_j\big)$
\State $r \leftarrow \operatorname{unit}\Big(\sum_{t\le K}w^{r}_t\,\tilde{x}_t
  \big/ \sum_{t\le K}w^{r}_t\Big)$, with
  $w^{r}_t=\exp\!\big(-(s^1_t-\min_{u\le K}s^1_u)/\gamma_r\big)$
\State $c'' \leftarrow (1-\omega_s)c'+\omega_s r$
\State \Return $(a'_j)_j$, $c'$, $c''$
\end{algorithmic}
\end{algorithm}

\noindent\textbf{Visual centre (lines~1--3).}
VIDE first scores the clip embeddings $\tilde{x}_t$ using the textual reference $\mu_n$ and the aggregation $\operatorname{head}(L_t)=\sum_{j=1}^{M}\pi_{t,j}L_{t,j}$ of PRISM (\cref{sec:setting}). It uses the initial scores $s^0_t$ to assign greater weight to clips that are more likely to be normal. The weight of clip $t$ is $w_t\propto\exp\!\big(-(s^0_t-\min_u s^0_u)/\gamma\big)$, normalised so that $\sum_t w_t=1$. A smaller temperature $\gamma$ concentrates the weights on the clips with the lowest initial scores. The normalised weighted mean $\hat{o}=\operatorname{unit}\big(\sum_t w_t\,\tilde{x}_t\big)$ estimates the normal centre $o$ defined in \cref{sec:discrepancies}.

\noindent\textbf{Reference correction and anomaly directions (lines~4--5).}
VIDE applies the soft projector $P$ to the displacement from $\mu_n$ to $\hat{o}$ and scales the correction by $\eta\in[0,1]$, giving $c'=\mu_n+\eta P(\hat{o}-\mu_n)$. The soft projector $P$ attenuates the components of $\hat{o}-\mu_n$ along directions with low variance among the prompt embeddings, which whitening would scale by up to $\lambda^{-1/2}$; this limits their influence on the recomputed anomaly directions. VIDE recomputes the anomaly directions as $a'_j=\operatorname{unit}\big(W(\mu_j-c')\big)$ and uses them with $c'$ to obtain the updated scores $s^1_t$.

\noindent\textbf{Scene anchor and scoring reference (lines~6--7).}
The scene anchor $r$ is the normalised weighted mean of the first $\min(K,T)$ clip embeddings. These weights favour clips with lower updated scores $s^1_t$, with their concentration controlled by $\gamma_r$. The scoring reference combines $c'$ with this anchor using weight $\omega_s$, $c''=(1-\omega_s)c'+\omega_s r$. The anchor enters only through $c''$, which adjusts the offsets of the unscaled logits $(\tilde{x}_t-c'')^\top a'_j$; the directions $a'_j$ are computed around $c'$, which avoids amplifying the anchor's components along directions with low prompt variance during whitening.

\vspace{-.5em}
\subsection{Axis calibration}
\label{sec:calibration}

PRISM aggregates the category logits using softmax weights. The analysis in \cref{sec:discrepancies} shows that the responses vary in dispersion and that text-derived directions can be poorly aligned with anomalous clips. VIDE rescales the logits, penalises strong responses across categories, and adjusts the logits using visual directions estimated from high-scoring clips. We apply \cref{alg:calibration} separately to each prompt view and omit the view index below. An overbar denotes the mean across clips, so $\bar{L}_j=\frac{1}{T}\sum_t L_{t,j}$.

\begin{algorithm}[t]
\caption{Axis calibration for one prompt view.}
\label{alg:calibration}
\fontsize{8}{9}\selectfont
\begin{algorithmic}[1]
\Require Clip embeddings $\tilde{x}_t$, directions $a'_j$, references $c'$ and
$c''$, scales $d_j$ (\cref{eq:scale}), score $S$ (\cref{eq:penalised}), and
parameters $\rho$, $\sigma_s$, $n_{\min}$, $\omega_m$
\Ensure Clip scores for the current view
\State $L_{t,j}\leftarrow(\tilde{x}_t-c'')^\top a'_j/d_j$
\State $s\leftarrow S(L)$, then \Return $s$ if $T<2n_{\min}$
\State $\mathcal{A}\leftarrow$ the $\max(n_{\min},\lceil\rho T\rceil)$ clips with the highest $g_t$, where $g$ is $s$ smoothed by a Gaussian filter of standard deviation $\sigma_s$
\State Assign each $t\in\mathcal{A}$ to the direction $j$ with the largest standardised logit $z_{t,j}$, giving sets $\mathcal{A}_j$
\State $L^*\leftarrow L$
\For{each direction $j$ with $|\mathcal{A}_j|\ge n_{\min}$}
  \State $u_j\leftarrow\operatorname{unit}\big(
    \frac{1}{|\mathcal{A}_j|}\sum_{t\in\mathcal{A}_j}\tilde{x}_t-c'\big)$,
    \quad $h_t\leftarrow(\tilde{x}_t-c')^\top u_j$
  \State $L^*_{t,j}\leftarrow(1-\omega_m)L_{t,j}
    +\omega_m\big(\bar{L}_j+\zeta_j(h_t-\bar{h})\big)$, with $\zeta_j=\operatorname{std}_t(L_{t,j})/\operatorname{std}_t(h_t)$
\EndFor
\State \Return $S(L^*)_t+b$ for each clip $t$, with $b=\max_t s_t-\max_t S(L^*)_t$
\end{algorithmic}
\end{algorithm}

\noindent\textbf{Logit scaling (line~1).}
We compute a text margin for each category using its original PRISM direction $a^0_j=\operatorname{unit}\big(W(\mu_j-\mu_n)\big)$. The margin $m_j=\max\big\{(\mu_j-\mu_n)^\top a^0_j,10^{-6}\big\}$ measures the separation between the normal and anomalous prompt means along $a^0_j$, subject to a positive lower bound. Given the mean margin $\bar{m}=\frac{1}{M}\sum_j m_j$ and exponent $p\ge 0$, the category scale is
\begin{equation}
  d_j=\left(\frac{m_j}{\bar{m}}\right)^p.
  \label{eq:scale}
\end{equation}
VIDE divides each projection $(\tilde{x}_t-c'')^\top a'_j$ by $d_j$: the logit of category $j$ is expressed in units of $d_j$, a scale given by the text. The scales are computed from the text statistics of each view. Setting $p=0$ gives $d_j=1$ for all categories.

\noindent\textbf{Penalty for broad category responses (line~2).}
A clip can receive a high anomaly score by responding strongly to several categories. To penalise such responses, VIDE subtracts the mean of the $q$ runner-up logits, scaled by $\beta$, from the softmax-weighted score. Let $L_{t,(i)}$ denote the $i$-th largest logit of clip $t$. The penalised score is
\begin{equation}
  S(L)_t
  = \sum_{j=1}^{M}\pi_{t,j}L_{t,j}
    - \frac{\beta}{q}\sum_{i=2}^{q+1} L_{t,(i)},
  \label{eq:penalised}
\end{equation}
where $\pi_t=\operatorname{softmax}(L_t/\tau)$, $q$ is the number of runner-up logits, and $\beta$ controls the penalty. For $\beta=0$, $S$ reduces to the PRISM aggregation.

\noindent\textbf{Adjustment using visual anomaly modes (lines~3--10).}
VIDE smooths the scores $s_t=S(L)_t$ with a Gaussian filter of standard
deviation $\sigma_s$ and selects the $\max(n_{\min},\lceil\rho T\rceil)$
highest-scoring clips. Each selected clip is assigned to the direction with
its largest standardised logit $z_{t,j}=(L_{t,j}-\bar{L}_j)/\operatorname{std}_t(L_{t,j})$. For a direction $j$ with at least
$n_{\min}$ assigned clips, VIDE computes a visual direction $u_j$ from
$c'$ toward the mean embedding of those clips. It projects each clip onto
this direction as $h_t=(\tilde{x}_t-c')^\top u_j$ and combines the resulting
response with the original logit, with mixing weight $\omega_m$ (line~8).
The factor $\zeta_j$ matches the standard deviation of the projection scores
to that of the original logits, and the offset $\bar{L}_j$ matches their mean
before mixing. For directions with fewer than $n_{\min}$ assigned clips, or whose
projection $h_t$ is constant, VIDE sets $L^*_{t,j}=L_{t,j}$. The
standardised logit of a category whose logits are constant is zero.
VIDE returns the scores $s_t$ if the video has fewer than $2n_{\min}$ clips
(line~2) or if no direction has enough assigned clips. The pooled metrics
compare scores across videos, so when a visual mode is used, VIDE preserves
the maximum score of the video before adjustment. It applies a shift
$b=\max_t s_t-\max_t S(L^*)_t$, yielding $S(L^*)_t+b$ as the score for clip $t$ (line~10).

\subsection{Multi-view aggregation}
\label{sec:views}

The analysis in \cref{sec:discrepancies} shows that anomaly scores are sensitive to prompt selection and that averaging scores across prompt subsets reduces this variability. The pseudo-code is given in \cref{alg:views} (\hyperref[sec:supp_neutral]{Appendix}).

\noindent\textbf{Temporal averaging.}
The window of $T_w$ clips around $t$ includes one more clip after $t$ than before it when $T_w$ is even, and is truncated at the video ends. This averaging reduces variations between neighbouring
clip embeddings. With $T_w=1$, $\tilde{x}_t=x_t$.

\noindent\textbf{Prompt views.}
The first view uses the full prompt bank. For each additional view $v$, we
sample, independently of the other views, up to $\nu$ anomaly prompts per
category without replacement, using seed $v-1$, and include all normal prompts. The same prompt sets are used
across videos, so their text statistics are computed once per encoder and view. VIDE applies \cref{alg:reference,alg:calibration} to each view and
averages the resulting clip scores to obtain $\bar{s}_t$. Relative to the full bank alone, the views raise the AUC on UCF-Crime by $0.22$ to $0.55$ points and the AP by $1.5$ to $2.3$ points on average over $20$ seed sets, all of which improve on the full bank alone (\hyperref[sec:supp_ablation]{Appendix}).

\noindent\textbf{Aggregation across similar clips.}
VIDE caps the neighbour count $\kappa$ at $\max(0,T-2\delta-1)$ so that every
clip has at least $\kappa$ eligible neighbours; if the capped count is zero, the final score is
$\tilde{s}_t=\bar{s}_t$. For each clip $t$, it selects
the $\kappa$ clips with the highest cosine similarity among those more
than $\delta$ steps away. The final score in \cref{eq:final} combines
$\bar{s}_t$ with the mean score of these neighbours, weighted by $\omega_c$.
A clip can receive a higher score when it resembles high-scoring
clips elsewhere in the video. VIDE excludes clips within $\delta$ steps
because the final smoothing already combines scores from nearby
clips.

\section{Experiments}
\label{sec:experiments}

\subsection{Experimental setup}
\label{sec:setup}

Following PRISM~\cite{chenprism}, we evaluate VIDE on the test sets of UCF-Crime (290 videos), XD-Violence (800 videos), and MSAD (240 videos). We report frame-level AUC on all three datasets, and average precision (AP) on XD-Violence and MSAD, and on UCF-Crime in Tab.~\ref{tab:supp_controls}. On MSAD, we additionally compute AUC and AP on the subset of videos containing at least one annotated anomaly, denoted by $\mathrm{AUC}_\mathrm{A}$ and $\mathrm{AP}_\mathrm{A}$.
We extract one visual embedding every 16 frames with each of two frozen encoders, Qwen3-VL-Embedding-2B~\cite{li2026qwen3} and PE-Core L14-336~\cite{bolya2026perception}. We use the PRISM prompt bank with 20 prompts per category. VIDE scores each test video independently without access to its annotations.
We select the scoring head and fusion hyperparameters with a leave-one-dataset-out protocol: each target dataset receives the configuration that maximises the mean selection score across the labelled test sets of the other two datasets, without using its own labels. The selection score is AUC on UCF-Crime and MSAD and the mean of AUC and AP on XD-Violence (configurations in the \hyperref[sec:supp_hyper]{Appendix}). For reference, a single configuration shared by the three datasets and selected with the labels of the three test sets reaches an AUC $0.01$ to $0.39$ points below that of leave-one-dataset-out (\hyperref[sec:supp_hyper]{Appendix}), so selecting the constants without target labels does not cost performance here. This shared configuration is not used for any result of the paper, all of which rely on leave-one-dataset-out. Each operation is applied on the three datasets with the same constants, except for two that lower the AUC when applied (the anchor on XD-Violence, $-0.47$; the views and neighbours on MSAD, $-0.78$; \hyperref[sec:supp_hyper]{Appendix}).



\begin{table}[t]
\centering

\caption{Comparison with state-of-the-art VAD methods on UCF-Crime and XD-Violence (AUC and AP in \%), grouped by setting and sorted by UCF-Crime AUC; PRISM and our results come last. $\dagger$: LLM-written prompts or generative VLM. Bold indicates the best result among training-free methods in each column.}
\label{tab:sota}
{\fontsize{6}{6.8}\selectfont\setlength{\tabcolsep}{1.2pt}\renewcommand{\arraystretch}{1.1}
\begin{tabular}{>{\raggedright\arraybackslash}p{1.06in}>{\raggedright\arraybackslash}p{0.68in}>{\centering\arraybackslash}p{0.4in}>{\centering\arraybackslash}p{0.33in}>{\centering\arraybackslash}p{0.33in}}
\toprule
Method & Setting & UCF-Crime & \multicolumn{2}{c}{XD-Violence}\\
\cmidrule(lr){3-3}\cmidrule(lr){4-5}
 & & AUC & AUC & AP\\
\midrule
\rowcolor{vtUns}Hasan et al.~\cite{hasan2016learning} & Unsupervised & 51.20 & -- & 50.32 \\
\rowcolor{vtUns}GODS~\cite{wang2019gods} & Unsupervised & 70.46 & -- & 61.56 \\
\rowcolor{vtUns}GCL~\cite{zaheer2022generative} & Unsupervised & 74.20 & -- & -- \\
\rowcolor{vtUns}DyAnNet~\cite{thakare2023dyannet} & Unsupervised & 79.76 & -- & -- \\
\midrule
\rowcolor{vtWeak}Sultani et al.~\cite{sultani2018real} & Weakly Supervised & 77.92 & -- & 73.20 \\
\rowcolor{vtWeak}SSRL~\cite{Li2022ScaleAwareSR} & Weakly Supervised & 87.43 & -- & -- \\
\rowcolor{vtWeak}CLIP-TSA~\cite{joo2023clip} & Weakly Supervised & 87.58 & -- & 82.17 \\
\rowcolor{vtWeak}VadCLIP~\cite{wu2024vadclip} & Weakly Supervised & 88.02 & -- & 84.51 \\
\rowcolor{vtWeak}Ex-VAD~\cite{huang2025ex} & Weakly Supervised & 88.02 & -- & 86.52 \\
\rowcolor{vtWeak}DSANet~\cite{yin2026learning} & Weakly Supervised & 89.44 & -- & 86.95 \\
\rowcolor{vtWeak}$\pi$-VAD~\cite{majhi2025just} & Weakly Supervised & 90.33 & -- & 85.37 \\
\rowcolor{vtWeak}GS-MoE~\cite{amicantonio2025mixture} & Weakly Supervised & 91.58 & 94.52 & 82.89 \\
\midrule
\rowcolor{vtFine}VADor~\cite{ztrk2023VADORRW} & Fine-tuning & 85.90 & -- & -- \\
\rowcolor{vtFine}VERA~\cite{ye2025vera} & Fine-tuning & 86.55 & 88.26 & 70.54 \\
\rowcolor{vtFine}Holmes-VAU~\cite{zhang2025holmes} & Fine-tuning & 88.96 & -- & 87.68 \\
\rowcolor{vtFine}Holmes-VAD~\cite{Zhang2024HolmesVADTU} & Fine-tuning & 89.51 & -- & 90.67 \\
\midrule
\rowcolor{vtTF}LLaVA-1.5~\cite{liu2024improved} & Training-free & 72.84 & 79.62 & 50.26 \\
\rowcolor{vtTF}Video-LLaMA2~\cite{cheng2024videollama} & Training-free & 74.42 & 80.21 & 53.57 \\
\rowcolor{vtTF}LAVAD~\cite{zanella2024harnessing} & Training-free & 80.28 & 85.36 & 62.01 \\
\rowcolor{vtTF}EventVAD~\cite{shao2025eventvad} & Training-free & 82.03 & 87.51 & 64.04 \\
\rowcolor{vtTF}MCANet~\cite{dev2024mcanet} & Training-free & 82.47 & 87.43 & 69.72 \\
\rowcolor{vtTF}SUVAD~\cite{gao2025suvad} & Training-free & 83.90 & -- & 70.10 \\
\rowcolor{vtTF}VADTree~\cite{li2026vadtree} & Training-free & 84.74 & 90.55 & 68.85 \\
\rowcolor{vtTF}Flashback~\cite{lee2025flashback} & Training-free & 87.29 & 90.54 & 75.13 \\
\rowcolor{vtTF}ASK-HINT$^\dagger$~\cite{Zou2025UnlockingVM} & Training-free & 89.83 & 90.31 & -- \\
\rowcolor{vtTF}NOVA$^\dagger$~\cite{Yin2026NOVANM} & Training-free & 89.86 & 95.07 & 84.82 \\
\rowcolor{vtFam}PRISM~\cite{chenprism} (PE)\vtbase & Training-free & 85.85 & 94.09 & 80.31 \\
\rowcolor{vtVIDE}VIDE (PE)\vtours & Training-free & 88.25 & 95.09 & 85.29 \\
\rowcolor{vtFam}PRISM~\cite{chenprism} (Qwen)\vtbase & Training-free & 88.85 & 93.42 & 77.48 \\
\rowcolor{vtVIDE}VIDE (Qwen)\vtours & Training-free & 89.67 & 93.83 & 81.50 \\
\rowcolor{vtVIDE}VIDE\vtours & Training-free & \textbf{90.32} & \textbf{95.94} & \textbf{86.84} \\
\bottomrule
\end{tabular}}
\end{table}

\subsection{State-of-the-art comparison and ablation}
\label{sec:sota}

\noindent\textbf{UCF-Crime (Tab.~\ref{tab:sota}).} VIDE achieves $90.32\%$ AUC, the highest among the compared training-free methods and $0.46$ points above NOVA. Compared with PRISM using the same encoder, VIDE improves AUC by $2.40$ points with PE-Core and $0.82$ points with Qwen. The fused result is $1.47$ points above the best published PRISM result.

\noindent\textbf{XD-Violence (Tab.~\ref{tab:sota}).} VIDE obtains the highest reported AUC ($95.94\%$) and the highest AP among the compared training-free methods ($86.84\%$). Relative to PRISM with PE-Core, the strongest PRISM configuration on this dataset, the gains are $1.85$ points in AUC and $6.53$ points in AP. The corresponding gains over NOVA are $0.87$ and $2.02$ points. In comparisons with PRISM using the same encoder, VIDE improves AP by $4.98$ points with PE-Core and $4.02$ points with Qwen.



\begin{table}[t]
\centering
\caption{Comparison on MSAD. $\mathrm{AUC}_\mathrm{A}$ and $\mathrm{AP}_\mathrm{A}$: videos that contain an anomaly. Grouped by setting and sorted by AUC; PRISM and our results come last. $^\dagger$Reproduced by us.}
\label{tab:msad}
{\fontsize{6}{6.8}\selectfont\setlength{\tabcolsep}{1.2pt}\renewcommand{\arraystretch}{1.0}
\begin{tabular}{>{\raggedright\arraybackslash}p{1.0in}>{\raggedright\arraybackslash}p{0.66in}>{\centering\arraybackslash}p{0.34in}>{\centering\arraybackslash}p{0.34in}>{\centering\arraybackslash}p{0.34in}>{\centering\arraybackslash}p{0.34in}}
\toprule
Method & Setting & AUC & AUC$_\mathrm{A}$ & AP & AP$_\mathrm{A}$\\
\midrule
\rowcolor{vtWeak}MGFN~\cite{Chen2022MGFNMG} & Weakly Supervised & 84.96 & -- & -- & -- \\
\rowcolor{vtWeak}UR-DMU~\cite{zhou2023dual} & Weakly Supervised & 85.78 & 67.95 & 67.35 & 75.30 \\
\rowcolor{vtWeak}RTFM~\cite{tian2021weakly} & Weakly Supervised & 86.65 & -- & -- & -- \\
\rowcolor{vtWeak}TEVAD~\cite{chen2023tevad} & Weakly Supervised & 86.82 & -- & -- & -- \\
\rowcolor{vtWeak}GS-MoE~\cite{amicantonio2025mixture} & Weakly Supervised & 87.72 & 69.54 & 68.26 & 76.68 \\
\rowcolor{vtWeak}$\pi$-VAD~\cite{majhi2025just} & Weakly Supervised & 88.68 & 71.25 & 71.26 & 77.86 \\
\midrule
\rowcolor{vtTF}VADTree~\cite{li2026vadtree} & Training-free & 89.32 & 67.85 & 71.41 & 75.49 \\
\rowcolor{vtTF}Flashback~\cite{lee2025flashback} & Training-free & 90.94 & 66.76 & 69.93 & 72.06 \\
\rowcolor{vtFam}PRISM~\cite{chenprism} (PE)$^\dagger$\vtbase & Training-free & 92.14 & 70.29 & 75.49 & 77.01 \\
\rowcolor{vtVIDE}VIDE (PE)\vtours & Training-free & 94.23 & 78.79 & 81.60 & 83.96 \\
\rowcolor{vtFam}PRISM~\cite{chenprism} (Qwen)\vtbase & Training-free & 91.84 & 69.58 & 75.48 & 76.50 \\
\rowcolor{vtVIDE}VIDE (Qwen)\vtours & Training-free & 94.75 & 79.72 & 82.22 & 83.47 \\
\rowcolor{vtVIDE}VIDE\vtours & Training-free & \textbf{95.33} & \textbf{81.13} & \textbf{84.75} & \textbf{85.87} \\
\bottomrule
\end{tabular}}
\end{table}


\noindent\textbf{MSAD (Tab.~\ref{tab:msad}).} VIDE outperforms the compared methods on all four metrics. Relative to PRISM with PE-Core (reproduced), its stronger encoder, VIDE increases AUC by $3.19$ points and AP by $9.26$ points on the full test set. Restricting evaluation to videos containing an anomaly, the gains are $10.84$ points in $\mathrm{AUC}_\mathrm{A}$ and $8.86$ points in $\mathrm{AP}_\mathrm{A}$.

\begin{table}[t]
\centering
\caption{Ablation of the operations (Op.), added cumulatively (last row of each block: complete VIDE). Op.~3 combines views, neighbours and the delay of the Qwen score. Baseline: PRISM.}
\label{tab:ablation}
{\fontsize{6}{6.8}\selectfont\setlength{\tabcolsep}{1.2pt}\renewcommand{\arraystretch}{1.1}
\begin{tabular}{>{\raggedright\arraybackslash}p{1.5in}>{\centering\arraybackslash}p{0.38in}>{\centering\arraybackslash}p{0.38in}>{\centering\arraybackslash}p{0.38in}>{\centering\arraybackslash}p{0.38in}}
\toprule
 & \multicolumn{2}{c}{XD-Violence} & \multicolumn{2}{c}{MSAD}\\
\cmidrule(lr){2-3}\cmidrule(lr){4-5}
Configuration & AUC & AP & AUC & AP\\
\midrule
\rowcolor{vtUns}\multicolumn{5}{l}{\textit{Qwen3-VL-Embedding}}\\
\rowcolor{vtFam}Baseline (PRISM, published) & 93.42 & 77.48 & 91.84 & 75.48 \\
\rowcolor{vtVIDE}\textbf{VIDE}\vtours &  &  &  &  \\
\rowcolor{vtVIDE!55}\hspace{1.0em}+ Visual reference (Op.~1) & 93.20 & 78.83 & 93.75 & 78.49 \\
\rowcolor{vtVIDE!55}\hspace{1.0em}+ Axis calibration (Op.~2) & 93.51 & 80.47 & 94.07 & 78.76 \\
\rowcolor{vtVIDE!55}\hspace{1.0em}+ Multi-view aggregation (Op.~3) & \textbf{93.83} & \textbf{81.50} & \textbf{94.75} & \textbf{82.22} \\
\midrule
\rowcolor{vtUns}\multicolumn{5}{l}{\textit{Fusion of the two encoders}}\\
\rowcolor{vtFam}Baseline (PRISM, published) & 94.09 & 80.31 & 91.84 & 75.48 \\
\rowcolor{vtVIDE}\textbf{VIDE}\vtours &  &  &  &  \\
\rowcolor{vtVIDE!55}\hspace{1.0em}+ Visual reference (Op.~1) & 95.43 & 84.31 & 95.03 & 84.46 \\
\rowcolor{vtVIDE!55}\hspace{1.0em}+ Axis calibration (Op.~2) & 95.80 & 86.45 & 95.06 & 84.39 \\
\rowcolor{vtVIDE!55}\hspace{1.0em}+ Multi-view aggregation (Op.~3) & \textbf{95.94} & \textbf{86.84} & \textbf{95.33} & \textbf{84.75} \\
\bottomrule
\end{tabular}}
\vspace{-.7em}
\end{table}

\noindent\textbf{Ablation (Tab.~\ref{tab:ablation}).} We add the three operations in sequence and compare each configuration with published PRISM results. The fusion block uses the stronger published PRISM encoder on each dataset as its baseline (only Qwen is published on MSAD). The first VIDE configuration introduces visual reference estimation together with the scoring head settings selected by leave-one-dataset-out. In the fusion block, this step also combines the two encoders. Adding axis calibration improves AUC in all four settings and raises AP on XD-Violence from $84.31\%$ to $86.45\%$ with fusion. Operation~3 further improves AUC in all four settings. For Qwen alone on MSAD, this operation delays the scores by three clips and increases AP from $78.76\%$ to $82.22\%$.

\noindent\textbf{Isolated gain of the operations (Tab.~\ref{tab:supp_controls}).} To isolate the operations, we evaluate PRISM with the head settings and fusion of VIDE, without retuning (\hyperref[sec:supp_controls]{Appendix}). This control differs from VIDE only by the operations, which add $3.57$, $2.86$ and $2.17$ AP points on UCF-Crime, XD-Violence and MSAD; the paired bootstrap interval of the difference in selection score over the videos excludes $0$ on the three datasets.

\begin{table}[t]
\centering
\caption{AP (\%) of the fusion. PRISM published: its stronger published encoder (Qwen alone on MSAD; no AP on UCF-Crime). Control: PRISM with the head settings and fusion of VIDE. Gain: VIDE minus Control, in points.}
\label{tab:supp_controls}
{\fontsize{6}{6.8}\selectfont\setlength{\tabcolsep}{1.2pt}\renewcommand{\arraystretch}{1.1}
\begin{tabular}{>{\raggedright\arraybackslash}p{0.72in}>{\centering\arraybackslash}p{0.55in}>{\centering\arraybackslash}p{0.55in}>{\centering\arraybackslash}p{0.55in}>{\centering\arraybackslash}p{0.55in}}
\toprule
 & PRISM published & Control & VIDE & Gain\\
\midrule
UCF-Crime & -- & 39.75 & \textbf{43.32} & +3.57 \\
XD-Violence & 80.31 & 83.98 & \textbf{86.84} & +2.86 \\
MSAD & 75.48 & 82.59 & \textbf{84.75} & +2.17 \\
\bottomrule
\end{tabular}}
\end{table}

\noindent\textbf{Leave-one-out (Tab.~\ref{tab:supp_removal}).} Removing one operation at a time from the complete VIDE, the other two being kept, lowers the selection score in all twelve cases, by $0.06$ to $1.25$ points. The largest losses are those of Op.~2 on XD-Violence ($-0.95$ with Qwen, $-1.25$ with the fusion) and of Op.~3 on MSAD with Qwen ($-0.69$). Op.~1 has the smallest loss in three of the four blocks ($0.10$ to $0.36$): the gain of the first step of Tab.~\ref{tab:ablation} should therefore not be read as the effect of Op.~1 alone.

\begin{table}[t]
\centering
\caption{Loss of the selection score (mean of AUC and AP on XD-Violence, AUC on MSAD), in points, when one operation is removed from the complete VIDE.}
\label{tab:supp_removal}
{\fontsize{6}{6.8}\selectfont\setlength{\tabcolsep}{1.2pt}\renewcommand{\arraystretch}{1.1}
\begin{tabular}{>{\raggedright\arraybackslash}p{1.5in}>{\centering\arraybackslash}p{0.38in}>{\centering\arraybackslash}p{0.38in}>{\centering\arraybackslash}p{0.38in}>{\centering\arraybackslash}p{0.38in}}
\toprule
 & \multicolumn{2}{c}{XD-Violence} & \multicolumn{2}{c}{MSAD}\\
\cmidrule(lr){2-3}\cmidrule(lr){4-5}
Configuration & Qwen & Fusion & Qwen & Fusion\\
\midrule
Without Visual reference (Op.~1) & $-$0.36 & $-$0.15 & $-$0.10 & $-$0.11 \\
Without Axis calibration (Op.~2) & $-$0.95 & $-$1.25 & $-$0.33 & $-$0.06 \\
Without Multi-view aggregation (Op.~3) & $-$0.68 & $-$0.26 & $-$0.69 & $-$0.27 \\
\bottomrule
\end{tabular}}
\end{table}


\begin{figure}[t]
\centering
\includegraphics[width=\columnwidth]{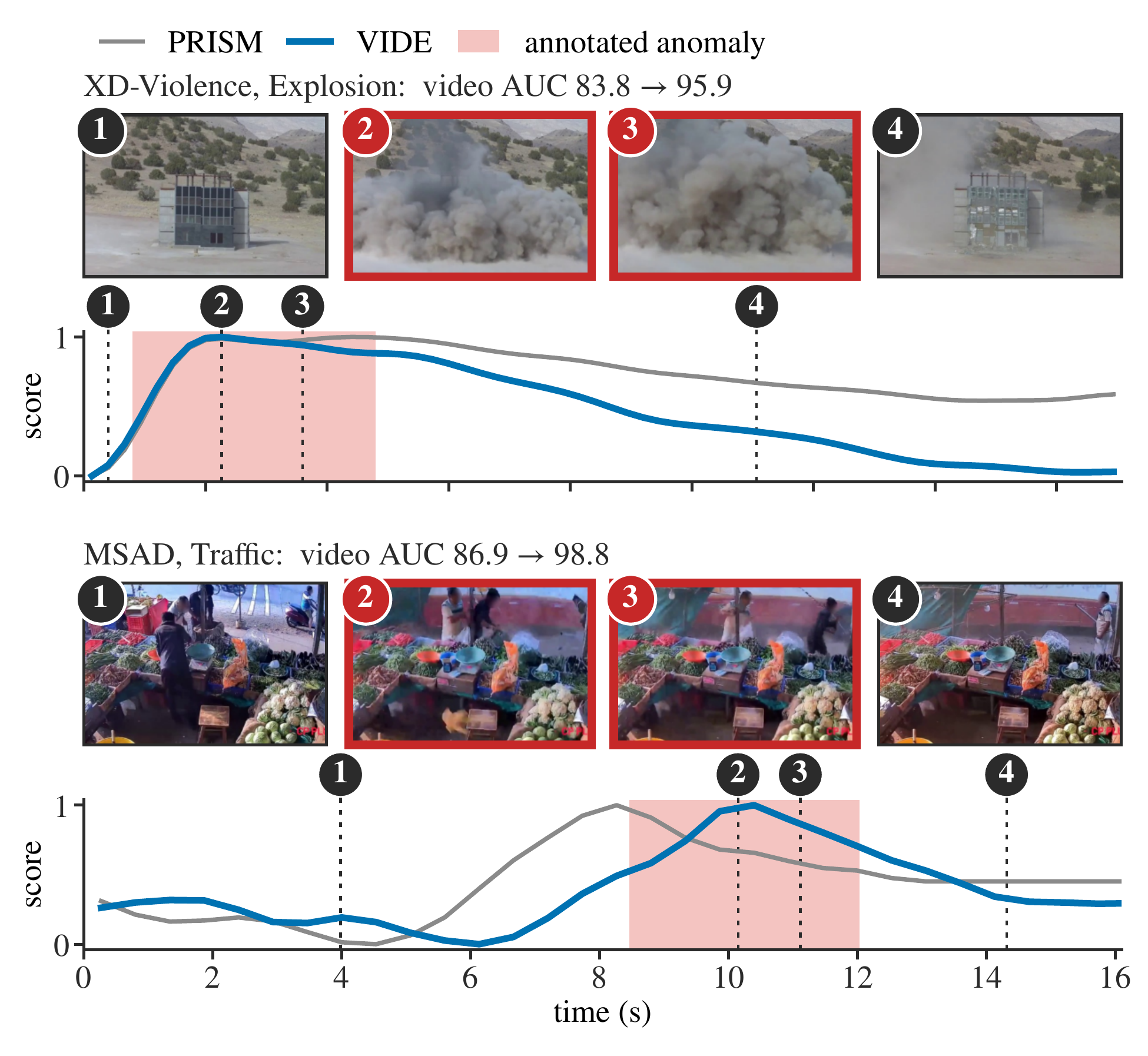}
\vspace{-2.em}
\caption{Qualitative results on XD-Violence and MSAD. Each key frame corresponds to the dotted line of the same number on the per-clip, min--max normalised scores of PRISM (grey) and VIDE (blue), with the same smoothing for both; red marks the annotated anomaly, and the title gives the AUC of the video, PRISM $\rightarrow$ VIDE. VIDE's score rises later on MSAD and falls faster after the explosion on XD-Violence.}
\label{fig:qualitative}
\vspace{-.2em}
\end{figure}

\vspace{-.5em}
\subsection{Qualitative results and efficiency}
\label{sec:qualitative}

Fig.~\ref{fig:qualitative} compares VIDE with PRISM using PE-Core on XD-Violence and Qwen on MSAD. For each example, the scores are plotted per clip and we apply the same Gaussian smoothing to both score sequences, with $\sigma_f=2$ clips on XD-Violence and $\sigma_f=1$ clip on MSAD. In the XD-Violence example, VIDE's score declines more rapidly after the explosion. The mean normalised score after the annotated interval is $0.36$ for VIDE and $0.71$ for PRISM. In the MSAD example, PRISM's score peaks before the annotated interval, while VIDE's score peaks within it.
Across videos containing both normal and anomalous frames, the median gain in per-video AUC is $1.8$ points on XD-Violence and $8.8$ points on MSAD. VIDE improves AUC on $64\%$ and $67\%$ of these videos, respectively. To assess the visual normality estimate $\hat{o}$, we compare it with the mean embedding $o$ of the annotated normal clips. With Qwen embeddings, their median cosine similarity is $0.99$ across $454$ XD-Violence videos. The advantage of VIDE over PRISM persists across anomaly proportions from $5\%$ to $90\%$: the AUC gap remains positive on the three datasets, at least $+0.59$ points (\hyperref[sec:supp_proportion]{Appendix}). The \hyperref[sec:supp_qual]{Appendix} provides the per-video statistics, further qualitative examples, a failure case, and a joint t-SNE visualisation of the embeddings.



\vspace{1.em}
\noindent\textbf{Efficiency.} We benchmark anomaly scoring on the $240$ MSAD test videos, with the same precomputed embeddings and videos for both methods, on a single thread of an AMD EPYC 7413 CPU. For each video we keep the best of $15$ runs; the time per clip is the sum of these times divided by the number of clips ($9{,}226$ with PE-Core, $7{,}790$ with Qwen), and the measurement, repeated three times, varies by less than $1\%$. With PE-Core, PRISM takes $4.1\,\mu\mathrm{s}$ per clip and VIDE $6.2\,\mu\mathrm{s}$; with Qwen, $7.1$ and $15.3\,\mu\mathrm{s}$, an overhead of $2.1$ to $8.2\,\mu\mathrm{s}$. With fusion, VIDE takes $0.77 \pm 0.28\,\mathrm{ms}$ per video (mean $\pm$ standard deviation over the $240$ videos), against $0.39\,\mathrm{ms}$ for the two PRISM heads. VIDE calls no LLM or generative model: end to end (extraction and inference), VIDE and PRISM take $1.1$ GPU-hours on these videos, against about $10$~\cite{li2026vadtree,chenprism,lee2025flashback} for LAVAD.



\section{Conclusion}
\label{sec:conclusion}

We presented VIDE, a training-free method for video anomaly detection that adapts the text-derived score to each test video. VIDE starts from the frozen embeddings of the video alone: it derives an estimate of the normal content of the video, which brings the textual reference closer to the observed scene. It then calibrates the anomaly responses, refines their directions using the highest-scoring clips, and averages the scores of several prompt views. It requires no labels on the target dataset, no training, and no generative model, and its overhead over the PRISM scoring head is $2.1$ to $8.2\,\mu\mathrm{s}$ per clip on MSAD. Used with each of the two encoders, VIDE improves PRISM. With both, it exceeds the best PRISM configuration by $6.53$ AP points on XD-Violence and $9.26$ on MSAD. VIDE obtains the highest AUC among the compared training-free methods on UCF-Crime, the highest among all compared methods on XD-Violence, and the best results on all four MSAD metrics.




\section*{Acknowledgments}
This work received partial support from Distech Controls Inc., the Natural Sciences and Engineering Research Council of Canada, the Digital Research Alliance of Canada, and MITACS.
\appendix
\clearpage
\setcounter{page}{1}
\maketitlesupplementary

\makeatletter\newcommand{\suppcaptionof}[2]{\def\@captype{#1}\caption{#2}}\makeatother

\begin{figure*}[t]
\centering
\includegraphics[width=\textwidth]{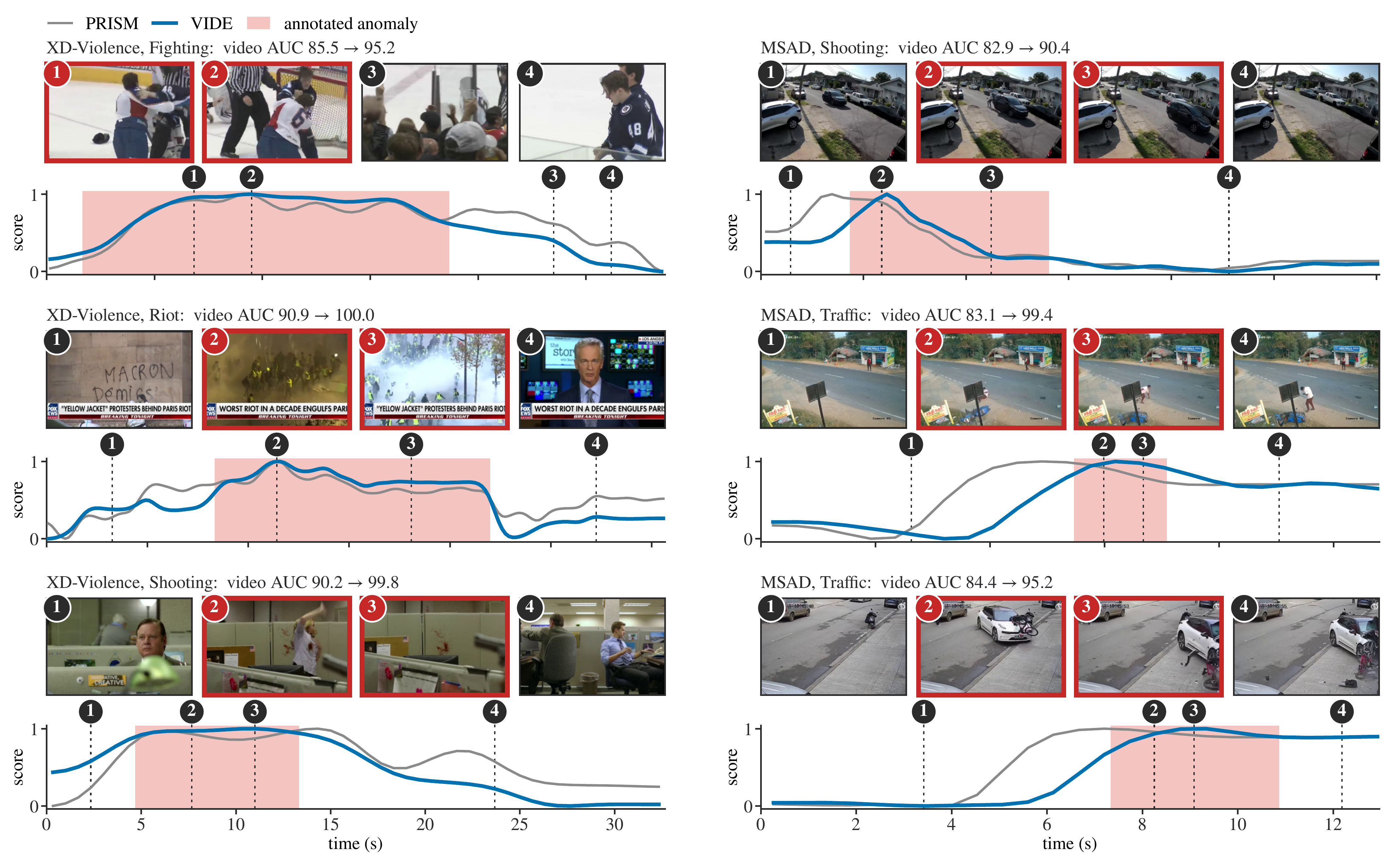}
\caption{Six further examples, three on XD-Violence (left column) and three on MSAD (right column). Same format as
\cref{fig:qualitative}: each key frame corresponds to the dotted line of the same number on the per-clip, min--max normalised
scores of PRISM (grey) and of VIDE (blue), with the same smoothing for both; red marks the annotated anomaly, and the
title gives the AUC of the video, PRISM $\rightarrow$ VIDE.}
\label{fig:supp_quartiles}
\end{figure*}

\noindent This supplementary material complements the paper, with the notation of the main paper, as follows:

\begin{itemize}
    \item Neutral cases and inputs of the algorithms (\cref{sec:supp_neutral})
    \item Hyperparameters (\cref{sec:supp_hyper})
    \item Leave-one-out ablation (\cref{sec:supp_ablation})
    \item Isolated gain of the operations (\cref{sec:supp_controls})
    \item Delay of the Qwen score (\cref{sec:supp_delay})
    \item Relation to NOVA (\cref{sec:supp_nova})
    \item Additional qualitative results (\cref{sec:supp_qual})
    \item Robustness to the proportion of anomalies (\cref{sec:supp_proportion})
\end{itemize}


\section{Neutral cases and inputs of the algorithms}
\label{sec:supp_neutral}

\noindent\textbf{Neutral cases of the three operations.}
Each operation has a neutral value of its parameters, for which it leaves the head unchanged. These values make each
operation removable on its own. With $\eta = 0$ and $\omega_s = 0$, Operation~1 gives $c' = c'' = \mu_n$ and $a'_j = a_j$:
the text reference and the axes are unchanged. With $p = \beta = \omega_m = 0$, Operation~2 gives $d_j = 1$, $S$ is the
score of the head, $L^* = L$ and $b = 0$. With $T_w = 1$, $V = 1$, $\kappa = 0$ and $\Delta = 0$, Operation~3 returns the
output of \cref{alg:calibration} on the whole bank of prompts, without delay of the Qwen score. When the three operations
are neutral, VIDE is the head of PRISM: PRISM is the special case of VIDE in which the three operations are neutral, the
delay of Operation~3 being zero ($\Delta = 0$).

\noindent\textbf{Inputs of the algorithms.}
\cref{alg:reference} takes the $T$ clips $\tilde{x}_t \in \mathbb{R}^D$ of the video, of unit norm after the temporal
averaging, and the text statistics of the view; its constants $\gamma$, $\eta$, $K$, $\gamma_r$ and $\omega_s$ are defined
where they are used. \cref{alg:views} takes the $T$ clips $x_t$ of the video, of unit norm, and the bank of prompts; its
constants are the length $T_w$ of the temporal window, the number of views $V$, the number $\nu$ of anomaly prompts per
category in a subset, the number $\kappa$ of neighbours, the exclusion radius $\delta$ and the weight $\omega_c$ of the
mixing. It reads only the video and the text: no label and no other video.

\begin{algorithm}[t]
\caption{Multi-view aggregation.}
\label{alg:views}
\fontsize{8}{9}\selectfont
\begin{algorithmic}[1]
\Require Clip embeddings $x_1,\dots,x_T$, prompt bank,
and parameters $V$, $\nu$, $T_w$, $\kappa$, $\delta$, $\omega_c$
\Ensure Clip scores $\tilde{s}_1,\dots,\tilde{s}_T$
\State Compute $\tilde{x}_t$ as the unit-normalised mean of the embeddings
  in the window of $T_w$ clips centred at $t$
\For{$v=1,\dots,V$}
  \State Use the text statistics of view $v$
  \State $(a'_{v,j},c'_v,c''_v)\leftarrow$ \cref{alg:reference} using
    $\tilde{x}$ and view $v$
  \State $s^{(v)}\leftarrow$ \cref{alg:calibration} using $\tilde{x}$,
    $a'_{v,j}$, $c'_v$, $c''_v$, and view $v$
\EndFor
\State $\bar{s}_t\leftarrow \frac{1}{V}\sum_{v=1}^{V}s^{(v)}_t$
\State $\kappa\leftarrow\min\{\kappa,\max(0,T-2\delta-1)\}$, then
  \Return $\bar{s}$ if $\kappa=0$
\State $\mathcal{N}_t\leftarrow$ the $\kappa$ clips $u$ with the largest
  $\tilde{x}_t^\top\tilde{x}_u$ among those satisfying $|t-u|>\delta$
\State $\tilde{s}_t\leftarrow(1-\omega_c)\bar{s}_t+
  \frac{\omega_c}{\kappa}\sum_{u\in\mathcal{N}_t}\bar{s}_u$
\State \Return $\tilde{s}$
\end{algorithmic}
\end{algorithm}

\begin{table}[t]
\centering
\caption{Hyperparameters of the head and of the fusion, and delay of Operation~3, for each target dataset (PE: PE-Core; Q: Qwen).
The values of a target dataset are selected by leave-one-dataset-out on the two other datasets (\cref{tab:supp_lodo}), without any label of the target dataset.
$\sigma$ and $\sigma_f$ are smoothing widths, in clips. $\phi$ and $\sigma_f$ apply to the fusion of both encoders;
$\Delta$ delays the Qwen score (Operation~3). The configuration common to the three datasets is given in
\cref{tab:supp_shared}.}
\label{tab:supp_hyper}
\small
\setlength{\tabcolsep}{5pt}
\begin{tabular}{llcccccc}
\toprule
 & & \multicolumn{2}{c}{UCF-Crime} & \multicolumn{2}{c}{XD-Violence} & \multicolumn{2}{c}{MSAD} \\
\cmidrule(lr){3-4}\cmidrule(lr){5-6}\cmidrule(lr){7-8}
 & & PE & Q & PE & Q & PE & Q \\
\midrule
Head   & $\tau$    & 0.21  & 0.5  & 0.1   & 0.21 & 0.05 & 0.14  \\
       & $\lambda$ & 0.003 & 0.01 & 0.003 & 0.03 & 0.3  & 0.001 \\
       & $\sigma$  & 4     & 8    & 2     & 4    & 1    & 1     \\
\midrule
Op.~3  & $\Delta$  & \multicolumn{2}{c}{2}     & \multicolumn{2}{c}{2}     & \multicolumn{2}{c}{3} \\
\midrule
Fusion & $\phi$    & \multicolumn{2}{c}{1.189} & \multicolumn{2}{c}{0.532} & \multicolumn{2}{c}{1.723} \\
       & $\sigma_f$& \multicolumn{2}{c}{6}     & \multicolumn{2}{c}{2}     & \multicolumn{2}{c}{1} \\
\bottomrule
\end{tabular}
\end{table}

\begin{table}[t]
\centering
\caption{Configuration common to the three datasets (UCF-Crime, XD-Violence and MSAD): the same values on each of them.
It is selected with the labels of the three test sets, so it does not follow leave-one-dataset-out and is not used in the paper; it serves as a reference.
Notation as in \cref{tab:supp_hyper}.}
\label{tab:supp_shared}
\small
\setlength{\tabcolsep}{8pt}
\begin{tabular}{llcc}
\toprule
 & & PE & Q \\
\midrule
Head   & $\tau$    & 0.2   & 0.3  \\
       & $\lambda$ & 0.001 & 0.01 \\
       & $\sigma$  & 2     & 6    \\
\midrule
Op.~3  & $\Delta$  & \multicolumn{2}{c}{2}   \\
\midrule
Fusion & $\phi$    & \multicolumn{2}{c}{0.5} \\
       & $\sigma_f$& \multicolumn{2}{c}{2}   \\
\bottomrule
\end{tabular}
\end{table}

\section{Hyperparameters}
\label{sec:supp_hyper}

\begin{table}[t]
\centering
\caption{AUC (\%) of PE and of the fusion with the constants selected by leave-one-dataset-out (\cref{tab:supp_hyper}), used in the paper, and with the common
configuration (\cref{tab:supp_shared}), selected with the labels of the three test sets. Diff.\ is the common value minus the leave-one-dataset-out value.}
\label{tab:supp_shared_results}
\small
\setlength{\tabcolsep}{3pt}
\begin{tabular}{lccc}
\toprule
 & UCF-Crime & XD-Violence & MSAD \\
\midrule
\multicolumn{4}{l}{\textit{PE}} \\
\quad Leave-one-dataset-out & 88.25   & 95.09   & 94.23   \\
\quad Common      & 88.18   & 94.70   & 94.22   \\
\quad Diff.       & $-0.06$ & $-0.39$ & $-0.01$ \\
\midrule
\multicolumn{4}{l}{\textit{Fusion}} \\
\quad Leave-one-dataset-out & 90.32   & 95.94   & 95.33   \\
\quad Common      & 90.09   & 95.62   & 95.20   \\
\quad Diff.       & $-0.23$ & $-0.31$ & $-0.13$ \\
\bottomrule
\end{tabular}
\end{table}

\begin{table}[t]
\centering
\caption{Leave-one-dataset-out: the constants of a target dataset are selected on the two other datasets.}
\label{tab:supp_lodo}
\small
\setlength{\tabcolsep}{8pt}
\begin{tabular}{ll}
\toprule
Target dataset & Selection on \\
\midrule
UCF-Crime   & XD-Violence + MSAD \\
XD-Violence & UCF-Crime + MSAD \\
MSAD        & UCF-Crime + XD-Violence \\
\bottomrule
\end{tabular}
\end{table}

VIDE has no learned parameter. Its constants are those of the head of PRISM ($\tau$, $\lambda$, $\sigma$), those of the
three operations, defined in \cref{alg:reference,alg:calibration,alg:views}, and those of the fusion of the two
encoders. \cref{tab:supp_hyper} gives the values of the head, of the delay of Operation~3 and of the fusion for each target dataset.
They are selected by leave-one-dataset-out (\cref{tab:supp_lodo}): for a target dataset, we keep the single configuration that maximises
the mean of the criteria of the two other datasets, measured on their labelled test sets (AUC on UCF-Crime and MSAD~\cite{zhu2024advancing}, mean of the AUC and the AP
on XD-Violence). The target dataset takes no part in this selection, neither through its labels nor through its videos. The procedure is run
three times, once per target dataset, and the results of the paper are those of these three configurations. \cref{tab:supp_shared} gives a
configuration common to the three datasets, and \cref{tab:supp_shared_results} compares it with leave-one-dataset-out.

Each operation is applied on the three datasets. On XD-Violence, the anchor of Operation~1 is not applied; on MSAD, the
views and the neighbours of Operation~3 are not applied. Applying them would lower the AUC by $0.47$ and $0.78$ points,
respectively on XD-Violence and MSAD.

\noindent\textbf{Head.} For each target dataset and each encoder, the head ($\tau$, $\lambda$, $\sigma$) is selected by
leave-one-dataset-out (\cref{tab:supp_hyper}), without any label of the target dataset. At inference, VIDE reads no label
and processes each video from itself and from the text.

\noindent\textbf{Fusion.} The scores of the two encoders are fused by
\begin{equation}
  F_t = \phi\, s^{\mathrm{Qwen}}_{t-\Delta} + s^{\mathrm{PE}}_t,
  \label{eq:supp_fusion}
\end{equation}
each score being smoothed by a Gaussian filter of width $\sigma_f$ before the sum. The Qwen score is delayed by
$\Delta$ clips (Operation~3), and its first score fills the start of the video. The values of $\phi$, $\sigma_f$ and $\Delta$ are given
for each target dataset (\cref{tab:supp_hyper}) and for the common configuration (\cref{tab:supp_shared}).

\noindent\textbf{Common configuration (reference).} To measure what leave-one-dataset-out leaves aside, we also searched for a single
configuration of the head and of the fusion, applied to every encoder and every dataset (\cref{tab:supp_shared}). Unlike the constants of
\cref{tab:supp_hyper}, it is tuned with the labels of the three test sets, so it does not follow leave-one-dataset-out and we do not use it
in the paper. The operations and their constants are not modified. \cref{tab:supp_shared_results} gives its AUC and the difference with
leave-one-dataset-out, which indicates how much a selection using the labels of the target dataset could bring. We report the results of
leave-one-dataset-out, which uses no label of the target dataset.

\noindent\textbf{Constants of the operations.} Where a component is applied, its constants take the same values
everywhere. Operation~1: $\eta = 0.1$ and $\gamma = 0.05$ for the visual reference; $\omega_s = 0.2$, $K = 32$ and
$\gamma_r = 1$ for the anchor. On XD-Violence, the anchor is an exception: $\omega_s = 0$ ($K$ and $\gamma_r$ then have
no effect). Operation~2: $p = 1$ for the scale; $\beta = 0.25$ and $q = 3$ for the penalty; $\rho = 0.05$,
$\sigma_s = 3$, $n_{\min} = 3$ and $\omega_m = 0.5$ for the tilt. Operation~3: $T_w = 5$, $\kappa = 5$, $\delta = 3$ and
$\omega_c = 0.5$ for the temporal mean and the neighbours; $V = 6$ views of $\nu = 5$ prompts per category; and
$\Delta = 2$ for the delay of the Qwen score. On MSAD, the views and the neighbours are an exception: $V = 1$,
$T_w = 1$ and $\kappa = 0$; so is the delay, $\Delta = 3$.

\begin{figure*}[t]
\centering
\includegraphics[width=\textwidth]{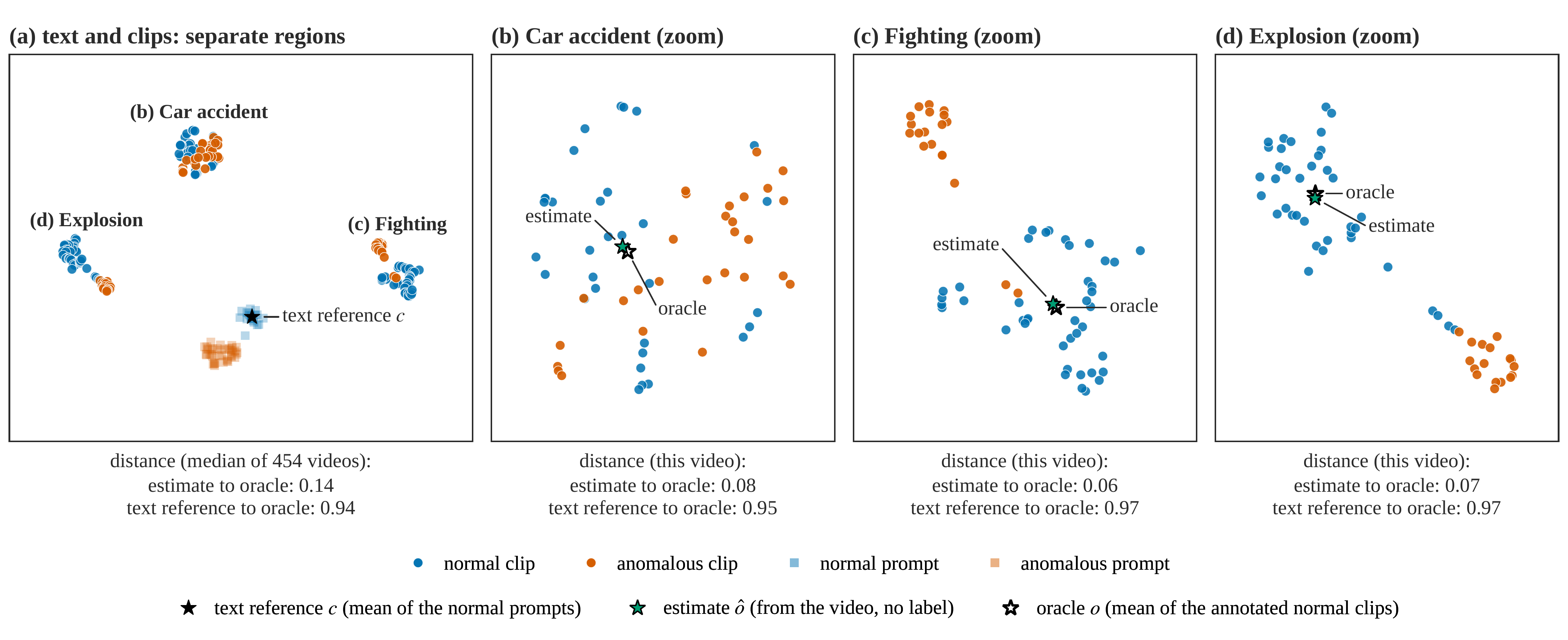}
\caption{Joint t-SNE (Qwen) of the prompts and of the clips of three XD-Violence videos. (a) The normal
prompts and the anomaly prompts (squares) form a text region, separate from those of the clips (circles); the text reference
$c$ (black star) lies in the text region. (b)--(d) Zoom on each video, with the true positions of the stars: the estimate
$\hat{o}$ (green star), computed from the video alone without any label, lies among the normal clips, next to the oracle $o$
(hollow star), the mean of the annotated normal clips. Under each panel: distance between the unit-norm embeddings of $\hat{o}$
and $o$. Median over the $454$ videos: $\cos(\hat{o}, o) = 0.991$.}\label{fig:supp_tsne}
\end{figure*}

\begin{figure}[t]
\centering
\includegraphics[width=\columnwidth]{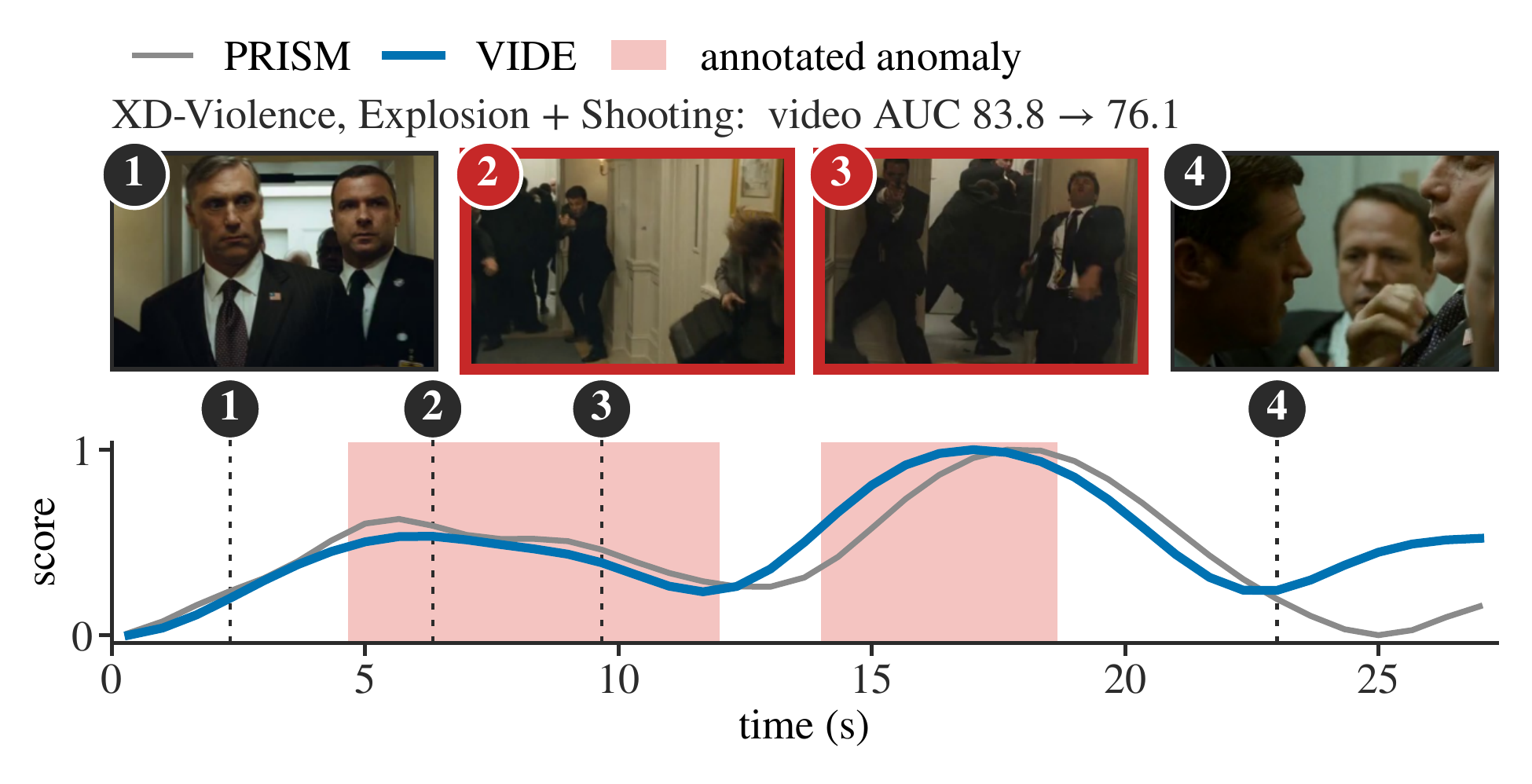}
\caption{Unfavourable case (XD-Violence, Salt.2010, 01:12:05--01:12:32), same format as \cref{fig:supp_quartiles}: the AUC of VIDE is lower than that of PRISM.}\label{fig:supp_unfav}
\end{figure}

\section{Leave-one-out ablation}
\label{sec:supp_ablation}

This section describes the leave-one-out ablation reported in \cref{tab:supp_removal} of the main paper, which complements the cumulative ablation of \cref{tab:ablation}. Starting from the complete VIDE, we remove each operation in turn, the other two being kept, on the same blocks as \cref{tab:ablation}, and we measure the loss on the selection criterion of each dataset (AUC on MSAD, mean of the AUC and the AP on XD-Violence). The two analyses answer different questions: \cref{tab:ablation} follows the path from PRISM to VIDE, and its first step also includes the selected head and the fusion, whereas the removal isolates the contribution of each operation within the complete VIDE. The losses and their discussion are given in the main paper.

\noindent\textbf{Logit scale.} The scale expresses the response of a clip in units of the text separation of its category, that is, the fraction of the displacement between the prompt means that the clip covers along the axis. The margins of a dataset differ by at most $27\,\%$ across categories, and the rank correlation between these margins and the dispersion of the logits within a video lies, in median, between $-0.09$ and $0.20$ depending on the dataset and the encoder: the scale is a correction derived from the text, which does not aim to equalise the visual dispersion. On XD-Violence, it improves the criterion of the fusion by $1.03$ points (PE: $+0.91$, Qwen: $+0.50$); on UCF-Crime, it improves the AUC of PE by $0.50$ points.

\noindent\textbf{Prompt views.} We compare the full prompt bank alone (one view) with the six views (the bank and five subsets of $\nu = 5$ prompts per category), the other operations and constants being fixed (\cref{tab:supp_views}). On UCF-Crime, the views of the paper (seeds $1$ to $5$) improve the AUC by $0.90$ points with PE, $0.22$ with Qwen and $0.39$ for the fusion, and the AP by $2.40$, $1.65$ and $2.23$ points. To measure the effect of the draw, we repeat the evaluation with $20$ other sets of $5$ seeds. Each of them improves the AUC and the AP of the three outputs. The mean gain in AUC is $0.55 \pm 0.23$ points with PE, $0.22 \pm 0.03$ with Qwen and $0.30 \pm 0.08$ for the fusion (standard deviation), and the mean gain in AP is $2.31 \pm 0.42$, $1.52 \pm 0.23$ and $1.96 \pm 0.32$. With PE, the seeds of the paper are more favourable than the average.

\begin{table}[t]
\centering
\caption{UCF-Crime (\%). Full bank alone, views with the seeds of the paper ($1$ to $5$), and mean $\pm$ standard deviation over $20$ other sets of $5$ seeds, the other operations and constants being fixed.}
\label{tab:supp_views}
\small
\setlength{\tabcolsep}{4pt}
\begin{tabular}{lccc}
\toprule
 & Bank alone & Paper seeds & Other seeds \\
\midrule
\multicolumn{4}{l}{\textit{AUC}} \\
\quad Qwen   & $89.45$ & $89.67$ & $89.67 \pm 0.03$ \\
\quad PE     & $87.35$ & $88.25$ & $87.90 \pm 0.23$ \\
\quad Fusion & $89.93$ & $90.32$ & $90.23 \pm 0.08$ \\
\midrule
\multicolumn{4}{l}{\textit{AP}} \\
\quad Qwen   & $39.85$ & $41.50$ & $41.37 \pm 0.23$ \\
\quad PE     & $35.27$ & $37.67$ & $37.58 \pm 0.42$ \\
\quad Fusion & $41.09$ & $43.32$ & $43.05 \pm 0.32$ \\
\bottomrule
\end{tabular}
\end{table}

\section{Isolated gain of the operations}
\label{sec:supp_controls}

The comparison with published PRISM (\cref{tab:ablation}) mixes three changes: the operations of VIDE, the head settings selected by leave-one-dataset-out, and the fusion of the two encoders. PRISM publishes no fusion, and its baseline is its stronger encoder alone. To isolate the operations, we evaluate PRISM with the head ($\tau$, $\lambda$, $\sigma$) of VIDE on each encoder, that is, VIDE with the three operations in their neutral case (\cref{sec:supp_neutral}), and we fuse it with the same two encoders and the same $\phi$ and $\sigma_f$ as VIDE (\cref{tab:supp_hyper}). This control and VIDE then differ only by the operations. The fusion is not tuned for the control, and the difference is measured with a paired bootstrap over the videos.

With the head settings and the fusion held fixed, the operations add $3.57$, $2.86$ and $2.17$ AP points on UCF-Crime, XD-Violence and MSAD (\cref{tab:supp_controls}). The paired bootstrap interval, computed on the selection criterion of each dataset, excludes $0$ on the three datasets.

\section{Delay of the Qwen score}
\label{sec:supp_delay}

A Qwen embedding covers $7$ consecutive clips and is stored at the first of them, so the Qwen score is delayed by $\Delta$ clips (Operation~3). PRISM has no such delay. To separate its effect, we evaluate PRISM and VIDE on MSAD with Qwen alone, with and without the delay ($\Delta = 3$), PRISM being given the head of VIDE (\cref{tab:supp_delay}).

\begin{table}[t]
\centering
\caption{MSAD, Qwen alone (\%). PRISM is given the head of VIDE; the delay is $\Delta = 3$.}
\label{tab:supp_delay}
\small
\setlength{\tabcolsep}{6pt}
\begin{tabular}{lcccc}
\toprule
 & AUC & AP & $\mathrm{AUC}_\mathrm{A}$ & $\mathrm{AP}_\mathrm{A}$ \\
\midrule
PRISM published     & $91.84$ & $75.48$ & $69.58$ & $76.50$ \\
PRISM               & $93.75$ & $78.41$ & $75.47$ & $79.69$ \\
PRISM, $\Delta = 3$ & $94.40$ & $81.51$ & $78.90$ & $82.90$ \\
VIDE                & $94.07$ & $78.76$ & $75.94$ & $79.88$ \\
VIDE, $\Delta = 3$  & $94.75$ & $82.22$ & $79.72$ & $83.47$ \\
\bottomrule
\end{tabular}
\end{table}

The delay is not specific to VIDE. It improves PRISM by $0.65$ AUC points and $3.10$ AP points, and VIDE by $0.68$ and $3.46$. Over delays from $0$ to $6$, $\Delta = 3$ is the best delay of both methods for AUC and AP, which is the centre of a window of $7$ clips. At equal delay, VIDE exceeds PRISM by $0.35$ AUC points, $0.71$ AP points, $0.82$ $\mathrm{AUC}_\mathrm{A}$ points and $0.57$ $\mathrm{AP}_\mathrm{A}$ points.

\section{Relation to NOVA}
\label{sec:supp_nova}

VIDE and NOVA~\cite{Yin2026NOVANM}, a concurrent training-free method, both estimate a visual reference from the clips of each test video. Each uses, in one of its operations, a weighting by calm and a normalised moving average, but the construction and the use of the reference differ.

\noindent\textbf{NOVA: the anchor is one more prompt.} NOVA computes the weights $w^{\mathrm{N}}_t \propto \exp(-\hat{s}_t/\gamma_{\mathrm{N}})$ on its first $N_{\mathrm{warm}} = 32$ embeddings, where $\hat{s}_t$ is a preliminary anomaly score, and the anchor $a_{\mathrm{VNA}} = \operatorname{unit}\big(\sum_{t \le N_{\mathrm{warm}}} w^{\mathrm{N}}_t x_t\big)$. A normalised moving average of $5$ embeddings precedes the scoring. The anchor is added to the normal prompts: the score of each anomaly category is $\mathrm{sigmoid}\big((\alpha\, \xi^{+}_t - \xi^{-}_t)/\tau_{\mathrm{N}}\big)$, where $\xi^{-}_t$ is the mean of the best $50\,\%$ of the similarities between $x_t$ and $\{e^{-}_i\}_{i=1}^{20} \cup \{a_{\mathrm{VNA}}\}$, $\xi^{+}_t$ is the same quantity for the anomaly prompts, and $\alpha < 1$ is a scaling factor. With $20$ normal prompts, $a_{\mathrm{VNA}}$ is one term out of $21$ and weighs at most one tenth of $\xi^{-}_t$. The prompts do not change. The anchor is computed on the beginning of the video and then frozen, which NOVA lists as a limitation, and a single anchor is shared by all categories.

\noindent\textbf{VIDE: the reference is corrected and the axes are rebuilt.} Our weights (\cref{alg:reference}, line~2) have the same form, since the subtraction of $\min_u s^0_u$ cancels in the normalisation, but they are applied to the whole video, with no assumption on the position of the normal clips: $\hat{o} = \operatorname{unit}\big(\sum_t w_t \tilde{x}_t\big)$. The score of VIDE has no competition between normal and anomaly prompts, so $\hat{o}$ is added to no prompt. It corrects the text reference $c = \mu_n$ into $c'$, and the axes are rebuilt around $c'$ (\cref{alg:reference}, line~4):
\begin{equation*}
  c' = \mu_n + \eta\, P(\hat{o} - \mu_n), \qquad a'_j = \operatorname{unit}\big(W(\mu_j - c')\big).
\end{equation*}
This correction answers the offset $(o - c)^\top a_j$ of \cref{eq:decomposition}, measured in \cref{sec:discrepancies}. The soft projector $P$ keeps from the displacement only the directions in which the prompts vary, and $\eta = 0.1$ bounds its amplitude. The anomaly directions of each category thus change, video by video. The anchor $r$ ($K = 32$, \cref{alg:reference}) enters afterwards, through the scoring reference $c'' = (1 - \omega_s)\,c' + \omega_s\, r$, with $\omega_s = 0.2$ ($\omega_s = 0$ on XD-Violence). For one prompt view, the unscaled logit is
\begin{equation*}
  (\tilde{x}_t - c'')^\top a'_j = (\tilde{x}_t - c')^\top a'_j - \omega_s\, (r - c')^\top a'_j .
\end{equation*}
The last term does not depend on the clip: $r$ shifts the origin of each axis, and is not a prototype to which the clips are compared. The moving average opens Operation~3, which adds $V = 6$ views of $\nu = 5$ prompts, the neighbours from the same video and the delay $\Delta$ of the Qwen score ($T_w = 5$ and $\kappa = 5$; on MSAD, $T_w = 1$, $V = 1$ and $\kappa = 0$).

\noindent\textbf{What VIDE adds, without LLM.} Operation~2 (scale per axis, penalty for responses on several categories, visual anomaly modes), the views, the neighbours, the delay and the fusion of two encoders have no counterpart in NOVA. VIDE uses no LLM. Removing Operation~1, which contains the weighting by calm but also the correction of the reference and the reconstruction of the axes, costs $0.15$ points of the criterion on XD-Violence and $0.11$ on MSAD with the fusion (\cref{tab:supp_removal}): the weighting by calm is therefore not the main source of the performance of VIDE. In the ablation of NOVA, the main gain ($+3.17$ AUC on UCF-Crime, $+3.97$ AP on XD-Violence) comes from prompt banks written by GPT-5.4-mini with guidance specific to each dataset, and its results are averaged over five banks; the anchor alone brings $+1.37$ AUC on UCF-Crime and $+0.11$ AP on XD-Violence.

\noindent\textbf{Results.} VIDE obtains $90.32$ AUC on UCF-Crime and $95.94$ AUC and $86.84$ AP on XD-Violence, against $89.86$ and $95.07$ / $84.82$ for NOVA (\cref{tab:sota}).

VIDE and NOVA therefore differ in the use of the reference (a projected correction and a reconstruction of the axes), in Operations~2 and~3, in the fusion of two encoders and in the absence of LLM.

\section{Additional qualitative results}
\label{sec:supp_qual}

This section contains the per-video statistics (\cref{tab:supp_stats}), six further examples (\cref{fig:supp_quartiles}), a joint t-SNE of the texts and of the clips, with the estimate $\hat{o}$ next to the oracle $o$ (\cref{fig:supp_tsne}), and an unfavourable case (\cref{fig:supp_unfav}).

\noindent\textbf{Per-video statistics (\cref{tab:supp_stats}).} The gain of a video is the difference between its AUC
with VIDE and with PRISM (better encoder: PE on XD-Violence, Qwen on MSAD). It is defined only for the videos that
contain both normal and anomalous frames. The two methods are smoothed by the same Gaussian filter ($\sigma_f = 2$ clips
on XD-Violence, $1$ clip on MSAD). The median gain is positive on both datasets, and more than half of the videos are
improved.

\begin{table}[t]
\centering
\caption{Per-video AUC gain of VIDE over PRISM.}
\label{tab:supp_stats}
\small
\setlength{\tabcolsep}{3pt}
\begin{tabular}{lccc}
\toprule
 & Videos & Median gain (points) & Improved videos \\
\midrule
XD-Violence & 500 & $+1.8$ & 64\,\% \\
MSAD        & 120 & $+8.8$ & 67\,\% \\
\bottomrule
\end{tabular}
\end{table}

\noindent\textbf{Additional examples (\cref{fig:supp_quartiles}).} In these six videos, we observe that VIDE improves the AUC
of the video, by $+7.5$ to $+16.3$ points. They do not represent the distribution of the gains of
\cref{tab:supp_stats}; \cref{fig:supp_unfav} shows a case where VIDE is below PRISM.

On XD-Violence (left column), the two scores are close inside the annotated segment, and the difference appears outside
it. In Fighting (segment from $3$ to $37$\,s), the score of PRISM stays higher after the segment (mean normalised level
$0.53$, against $0.31$ for VIDE). In Riot, it exceeds $0.5$ about $15$\,s before the start of the segment, against about
$1$\,s for VIDE. In Shooting, it is maximal $1$\,s after the end of the segment, whereas the maximum of VIDE lies inside
the segment.

\begin{figure*}[t]
\centering
\begin{minipage}{\textwidth}\centering
\suppcaptionof{table}{Pooled AUC (\%) on windows with a controlled proportion of anomalies, mean over five draws. $n$: anomalous
videos that admit a window at this proportion (all the normal videos are kept). Gap: VIDE minus PRISM, in points.}\label{tab:supp_proportion}
\footnotesize
\setlength{\tabcolsep}{4pt}\renewcommand{\arraystretch}{0.92}
\begin{tabular}{lcccccccccccc}
\toprule
 & \multicolumn{4}{c}{XD-Violence} & \multicolumn{4}{c}{UCF-Crime} & \multicolumn{4}{c}{MSAD} \\
\cmidrule(lr){2-5}\cmidrule(lr){6-9}\cmidrule(lr){10-13}
Proportion (\%) & $n$ & PRISM & VIDE & Gap & $n$ & PRISM & VIDE & Gap & $n$ & PRISM & VIDE & Gap \\
\midrule
5 & 294 & 92.39 & 93.22 & $+0.84$ & 120 & 84.02 & 86.93 & $+2.92$ & 12 & 83.08 & 87.60 & $+4.52$ \\
10 & 287 & 93.24 & 94.52 & $+1.28$ & 125 & 86.74 & 89.27 & $+2.53$ & 15 & 87.82 & 90.75 & $+2.93$ \\
20 & 326 & 93.48 & 95.08 & $+1.59$ & 118 & 88.56 & 91.25 & $+2.69$ & 14 & 89.38 & 93.34 & $+3.96$ \\
30 & 326 & 94.16 & 95.71 & $+1.55$ & 106 & 90.13 & 92.59 & $+2.45$ & 13 & 92.44 & 95.42 & $+2.98$ \\
40 & 308 & 95.28 & 96.63 & $+1.36$ & 98 & 91.15 & 93.59 & $+2.44$ & 15 & 91.48 & 95.54 & $+4.06$ \\
50 & 293 & 95.79 & 97.09 & $+1.30$ & 82 & 92.32 & 94.70 & $+2.37$ & 15 & 92.81 & 96.41 & $+3.60$ \\
60 & 257 & 96.55 & 97.72 & $+1.17$ & 75 & 93.15 & 95.35 & $+2.20$ & 16 & 93.74 & 96.56 & $+2.82$ \\
70 & 233 & 97.30 & 98.27 & $+0.97$ & 71 & 93.95 & 95.98 & $+2.03$ & 19 & 95.08 & 97.30 & $+2.22$ \\
80 & 208 & 97.75 & 98.57 & $+0.82$ & 59 & 94.62 & 96.49 & $+1.87$ & 18 & 95.84 & 97.72 & $+1.89$ \\
90 & 182 & 98.38 & 98.98 & $+0.59$ & 51 & 95.03 & 96.85 & $+1.82$ & 20 & 96.22 & 97.78 & $+1.56$ \\
\bottomrule
\end{tabular}

\vspace{4pt}
\includegraphics[width=\textwidth]{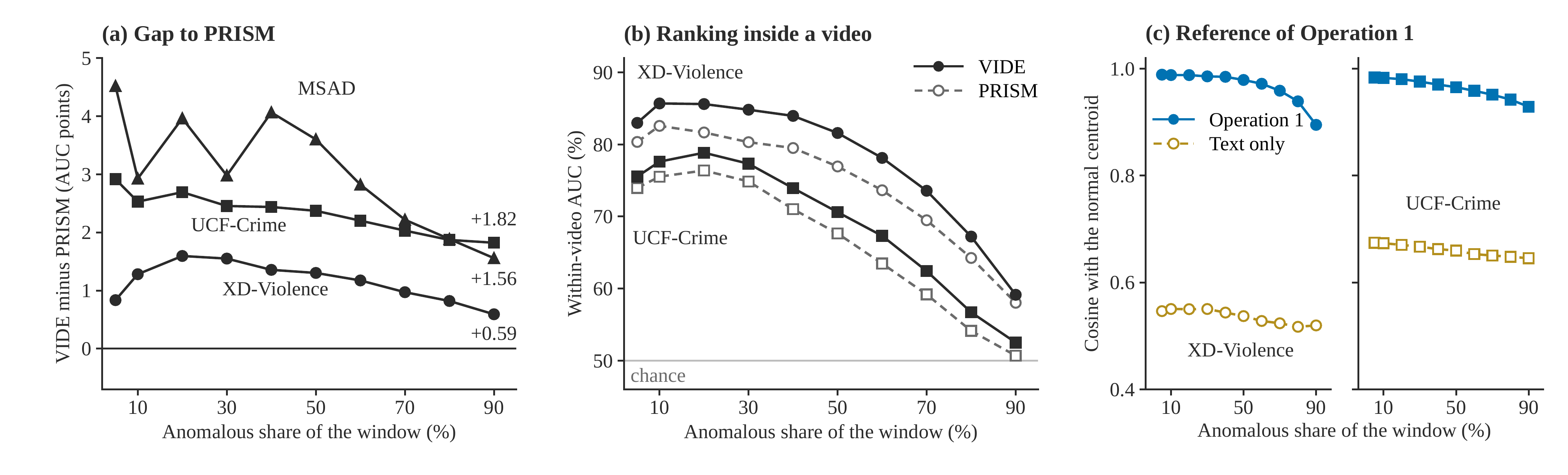}
\suppcaptionof{figure}{Robustness to the proportion of anomalies in the window: VIDE stays above PRISM up to $90\,\%$, with a
gap that decreases. (a) Pooled AUC gap between VIDE and PRISM, with the values at $90\,\%$. (b) Within-video AUC of VIDE
(solid) and of PRISM (dashed); the grey line is the chance level. (c) Cosine with the centroid of the normal clips of the
video of the reference estimated by Operation~1 (blue) and of the text reference alone (yellow, dashed). Circle:
XD-Violence; square: UCF-Crime; triangle: MSAD ($12$ to $20$ videos per proportion).}\label{fig:supp_proportion}
\end{minipage}
\end{figure*}

On MSAD (right column), the score of PRISM rises before the accident in all three videos. In the first row (Shooting),
it is already high at the start of the video and peaks $1$\,s before the segment; in the second (Traffic), it exceeds
$0.5$ about $5$\,s before the segment, and in the third (Traffic) about $2$\,s before. The score of VIDE rises later and
reaches its maximum inside the annotated segment.

\noindent\textbf{Joint t-SNE (\cref{fig:supp_tsne}).} The figure projects together the normal prompts, the anomaly prompts and
the clips of three XD-Violence videos of different categories (Car accident, Fighting, Explosion), with the Qwen encoder. The text and the clips occupy separate regions (a); the text reference $c$, the mean of the normal
prompts, lies in the text region.

The estimate $\hat{o}$, computed by Operation~1 from the clips of the video alone, without any label, lies among
the normal clips, next to the oracle $o$ (b)--(d). The oracle serves only as a measure: VIDE never uses it.

The distances written under the panels and the cosines, both computed on the embeddings and not on the projection, confirm
it: $\hat{o}$ is at a distance of $0.08$, $0.06$ and $0.07$ from $o$ ($\cos(\hat{o}, o) = 0.997$, $0.998$ and $0.998$). On the
$454$ annotated videos of XD-Violence, the median distance is $0.14$ ($\cos = 0.99$), and three quarters of the videos have
$\cos(\hat{o}, o) \geq 0.976$. The normal centre of a video is thus read in the video itself.

\noindent\textbf{Unfavourable case (\cref{fig:supp_unfav}).} On the XD-Violence video Salt.2010 (01:12:05--01:12:32),
the AUC goes from $83.8$ with PRISM to $76.1$ with VIDE. The video is short ($41$ clips, $27$\,s), and its anomaly, split
into two segments of $11$ and $7$ clips, occupies $44\%$ of it. In the first segment, the score of VIDE is slightly lower
than that of PRISM (mean normalised level $0.43$, against $0.49$); after the second, it rises back to $0.39$, whereas
that of PRISM falls to $0.11$.

Among the operations active on XD-Violence, the one that best explains the gap is the context of the PE encoder
(Operation~3: a window of $T_w = 5$ clips and $\kappa = 5$ neighbours). Removing it gives back $5.8$ points, i.e.\ $75\%$
of the gap (the AUC rises to $81.9$, the fusion weight being tuned again), whereas no other operation gives back more
than $3.9$ points. A plausible explanation lies in the length of the video: with $41$ clips, almost half of them
anomalous, the window and the neighbours of a normal clip often contain anomalous clips, which reduces the contrast
between the anomaly and the background. \cref{sec:supp_proportion} measures the effect of the proportion of anomalies.

\section{Robustness to the proportion of anomalies}
\label{sec:supp_proportion}

This section measures how far VIDE holds when the share of anomalies in a video increases (\cref{tab:supp_proportion},
\cref{fig:supp_proportion}).

\noindent\textbf{Motivation.} Three mechanisms of VIDE rely on the rarity of the anomaly: the visual centre $\hat{o}$ of
Operation~1, estimated from the calm clips, the anomaly modes of Operation~2 and the neighbours of Operation~3. When the
share of anomalous clips increases, $\hat{o}$ is contaminated and the modes may be wrong. The datasets do not allow us to
measure this. The median proportion of anomalies in the anomalous videos is $14\,\%$ on UCF-Crime, $35\,\%$ on XD-Violence
and $58\,\%$ on MSAD, and $7\,\%$, $35\,\%$ and $59\,\%$ of these videos, respectively, exceed $50\,\%$. We therefore vary
it within a single video, without retraining and without any additional tuning.

\noindent\textbf{Protocol.} For each anomalous video and each target proportion from $5\,\%$ to $90\,\%$ ($5$, $10$, $20$,
\ldots, $90$), we crop the video into a contiguous window that meets three conditions: its share of anomalies, measured as
in the protocol of the dataset (clips on XD-Violence, frames on UCF-Crime and MSAD), equals the target proportion to within
one point; it contains at least one anomaly; and it contains at least $24$ samples. The window is drawn at random among the
longest possible ones (at least $90\,\%$ of the maximal length), to keep as much context as possible. The Qwen embeddings
are cropped with it, so that none extends beyond the window. Normal videos are left unchanged. A video enters a proportion
only if it admits such a window; the number of videos, which varies, is given in \cref{tab:supp_proportion}. The models are
those of the main experiments, without any tuning. PRISM is evaluated with its better encoder (PE on XD-Violence, Qwen on
UCF-Crime and MSAD). Both methods are smoothed by the same Gaussian filter ($\sigma_f = 2$ clips on XD-Violence, $6$ on
UCF-Crime and $1$ on MSAD). The results are means over five draws. We report the pooled AUC of the protocol of each
dataset, and the within-video AUC, computed after centring the scores of each video on their own mean, which keeps only the
order of the clips inside a video.

\noindent\textbf{Gap to PRISM (\cref{tab:supp_proportion}, \cref{fig:supp_proportion}a).} VIDE is above PRISM at every
proportion, from $5\,\%$ to $90\,\%$, on the three datasets and in each of the five draws. The gap is not constant,
however. On XD-Violence, it goes from $+0.84$ points at $5\,\%$ to $+1.59$ at $20\,\%$, then decreases to $+0.59$ at
$90\,\%$. On UCF-Crime, it goes from $+2.92$ at $5\,\%$ to $+1.82$ at $90\,\%$, with $+2.37$ at $50\,\%$. On MSAD, it goes
from $+4.52$ to $+1.56$, but it rests on only $12$ to $20$ videos per proportion. The pooled AUC of each method increases
with the proportion, partly because the normal clips of the anomalous videos, the hardest to reject, become scarce. It
therefore does not measure robustness on its own, and we read the gap between the two methods together with the
within-video AUC.

\noindent\textbf{Loss of discrimination (\cref{fig:supp_proportion}b).} The within-video AUC falls with the proportion, for
VIDE as for PRISM. On XD-Violence, it goes for VIDE from $85.69$ at $10\,\%$ to $59.13$ at $90\,\%$, and for PRISM from
$82.58$ to $58.02$. On UCF-Crime, it goes from $77.61$ to $52.50$ for VIDE and from $75.53$ to $50.68$ for PRISM. At
$90\,\%$, both methods are below $60\,\%$, close to the chance level: few normal clips remain to compare with. VIDE stays
above PRISM at every proportion, by $+1.1$ to $+4.6$ points on XD-Violence and by $+1.6$ to $+3.8$ on UCF-Crime. MSAD, which
has only $12$ to $20$ videos per proportion, is shown neither here nor in (c).

\noindent\textbf{Reference of Operation~1 (\cref{fig:supp_proportion}c).} The cosine between the reference estimated by
Operation~1 ($\hat{o}$ on XD-Violence, the anchor on UCF-Crime) and the centroid of the normal clips of the video stays
above $0.95$ up to $70\,\%$: from $0.989$ at $5\,\%$ to $0.959$ at $70\,\%$ on XD-Violence, and from $0.984$ to $0.951$ on
UCF-Crime. It then decreases to $0.895$ and $0.929$ at $90\,\%$ (on $56$ and $19$ videos on average over the draws), but
stays well above that of the text reference alone ($0.52$ and $0.65$). Removing Operation~1 changes the pooled AUC of VIDE
by at most $0.6$ points at every proportion; on UCF-Crime, from $60\,\%$ onwards, this removal is slightly favourable (up
to $0.34$ points at $90\,\%$). The anomaly modes of Operation~2 and the neighbours of Operation~3 are not isolated: their
effect is included in the gap to PRISM. VIDE therefore keeps its advantage over PRISM up to $90\,\%$ of anomalies, but
this advantage shrinks as the proportion increases.

\clearpage
{
    \small
    \bibliographystyle{ieeenat_fullname}
    \bibliography{main}
}


\end{document}